\documentclass[conference,compsoc]{IEEEtran}
\usepackage{graphicx}
\graphicspath{ {./figures/} }
\usepackage{algorithm}
\usepackage{algpseudocode}
\usepackage{algorithmicx}
\usepackage{multirow}
\usepackage{algorithm,algpseudocode}
\usepackage{xcolor}
\usepackage{todonotes}
\usepackage{xcolor}
\usepackage{soul}
\usepackage{color}
\usepackage{balance}
\usepackage{amsmath}
\usepackage{booktabs}
\usepackage{hyperref}

\newcommand{\model}{\textbf{\textit{w}}}

\makeatletter
\newcommand{\newlineauthors}{%
\end{@IEEEauthorhalign}\hfill\mbox{}\par
\mbox{}\hfill\begin{@IEEEauthorhalign}
}
\makeatother
\presetkeys%
    {todonotes}%
    {backgroundcolor=green!40}{}
\begin{document}
%
\title{Label Inference Attacks against Node-level Vertical Federated GNNs}

\author{\IEEEauthorblockN{Marco Arazzi}
	\IEEEauthorblockA{University of Pavia\\
		Pavia, Italy \\
		marco.arazzi01@universitadipavia.it}
	\and
	\IEEEauthorblockN{Mauro Conti}
	\IEEEauthorblockA{University of Padua \& \\
		Delft University of Technology\\
		Padova, Italy \\
		mauro.conti@unipd.it}
	\and
	\IEEEauthorblockN{Stefanos Koffas}
	\IEEEauthorblockA{Delft University of Technology\\
		Delft, The Netherlands \\
		S.Koffas@tudelft.nl}
	\newlineauthors
	\IEEEauthorblockN{Marina Krcek}
	\IEEEauthorblockA{Delft University of Technology\\
		Delft, The Netherlands \\
		M.Krcek@tudelft.nl}
	\and
	\IEEEauthorblockN{Antonino Nocera}
	\IEEEauthorblockA{University of Pavia\\
		Pavia, Italy \\
		antonino.nocera@unipv.it}
	\and
	\IEEEauthorblockN{Stjepan Picek}
	\IEEEauthorblockA{Radboud University \& \\
		Delft University of Technology\\
		Nijmegen, The Netherlands \\
		stjepan.picek@ru.nl}
	\newlineauthors
	\IEEEauthorblockN{Jing Xu}
	\IEEEauthorblockA{Delft University of Technology\\
		Delft, The Netherlands \\
		J.Xu-8@tudelft.nl}
}

\maketitle

\begin{abstract}
Federated learning enables collaborative training of machine learning models by keeping the raw data of the involved workers private. Three of its main objectives are to improve the models' privacy, security, and scalability. Vertical Federated Learning (VFL) offers an efficient cross-silo setting where a few parties collaboratively train a model without sharing the same features. In such a scenario, classification labels are commonly considered sensitive information held exclusively by one (active) party, while other (passive) parties use only their local information. 
Recent works have uncovered important flaws of VFL, leading to possible label inference attacks under the assumption that the attacker has some, even limited, background knowledge on the relation between labels and data. 
In this work, we are the first (to the best of our knowledge) to investigate label inference attacks on VFL using a zero-background knowledge strategy.
To formulate our proposal, we focus on Graph Neural Networks (GNNs) as a target model for the underlying VFL. In particular, we refer to node classification tasks, which are widely studied, and GNNs have shown promising results.
Our proposed attack, BlindSage, provides impressive results in the experiments, achieving nearly 100\% accuracy in most cases. Even when the attacker has no information about the used architecture or the number of classes, the accuracy remains above 90\% in most instances.
Finally, we observe that well-known defenses cannot mitigate our attack without affecting the model's performance on the main classification task.
\end{abstract}



%

\section{Introduction}
\label{sec:introduction}

Federated Learning (FL) has emerged as a groundbreaking paradigm that enables collaborative training of machine learning models across decentralized data sources without sharing raw data~\cite{konevcny2016federated}. This distributed approach aims to provide data privacy, security, and scalability, making it particularly appealing for sensitive applications in various domains. 
FL can be categorized into Horizontal FL (HFL), Vertical FL (VFL), and Federated Transfer Learning (FTL).
In HFL, clients participating in FL share datasets of the same feature space but hold different examples. On the contrary, clients in VFL have the same data examples but different feature sets. In FTL, the clients' data differ in feature space and instances.
This work focuses on vertical data distribution as it is a realistic threat model, is widely used in the industry, and not yet well explored by the community~\cite{a-survey-on-federated-learning-systems-vision-hype-and-reality-for-data-privacy-protection}. 
Moreover, in VFL, the fact that the feature space is divided among involved clients already provides a more significant challenge than when data samples are distributed between parties.
Additionally, clients in VFL control only their local model, unlike HFL, where they control the entire global model, and this is compounded by potential variations in local model architectures across clients.
However, as VFL becomes more popular~\cite{federated-machine-learning-concept-and-applications,interpret-federated-learning-with-shapley-values}, so do the concerns about potential security vulnerabilities that might compromise the privacy of sensitive data. 
Our work focuses on VFL scenarios where active parties have access to the dataset labels for the main task (e.g., classification) while passive clients do not know them. 
We propose an advanced label inference attack to extract confidential information from the active clients, with the attacker being one of the passive parties.

Label inference attacks on VFL have already been introduced in the literature~\cite{fu2022label}. Such attacks require some background knowledge about the possible labels. In particular, recent works show that if attackers have access to a limited dataset with a subset of its labels, they can infer the private labels of the whole dataset~\cite{fu2022label}. 
In our work, we start from these results and propose a novel attack that does not require any background knowledge of the labels in the considered dataset or on the target global classification model. According to our solution, the attack can be mounted only by using the gradients returned by the server during the federated training, thus making our attack hard to detect, stealthy, and more realistic as it assumes a weaker attacker. Additionally, our attack can be applied offline using the saved history of the returned gradients so that the training costs are not increased. 
To evaluate the feasibility and performance of our proposal, we define the attack against a target federated learning model based on Graph Neural Networks (GNNs), which is an interesting research direction~\cite{fu2022label} and a relevant application for VFL~\cite{ni2021vertical,chen2020vertically}. 
More specifically, GNNs have recently gained remarkable attention due to their ability to handle graph data, which is common in many real-world problems. Such data can represent social networks, recommendation systems such as the in-house Pinterest's model named PinSage~\cite{ying2018graph}, and DNA sequences in bioinformatics~\cite{wu2020comprehensive}. 
Node classification is a popular and important task where GNNs are used with a very satisfactory performance~\cite{zhou2020graph, wu2020comprehensive, wu2022graph}. For example, GNNs have been used to detect fraudulent entities in a computer network~\cite{liu2020alleviating}. Other popular applications of the node classification task include detecting malicious users in social networks~\cite{liu2018heterogeneous}, classifying the function of proteins in the interactions~\cite{hamilton2017inductiverepr_graphsage_reddit}, and classifying the topic of documents based on hyperlink or citation graphs~\cite{kipf2017semigcn}. 
In many application contexts mentioned above, data are vertically partitioned among several heterogeneous platforms.
That is true, for instance, in the context of social networks~\cite{vosecky2009user,2015discovering}, where over the years, a number of research works have focused on studying unified user profiles by suitably exploiting and combining the information available in these independent systems~\cite{MultySNModel14,silvestri2015linking}.
Federated learning provides an important solution in this direction, allowing GNNs to be utilized for modeling graph-structured data across multiple participants to preserve data locality. 
GNNs can be used in VFL for real-world node classification tasks~\cite{chen2022graphfraudster}. Consider, for instance, the practical example of money lending, where banks want to avoid lending money to users with low credit ratings.
This procedure requires a reliable evaluation agent to assess the target users among financial parties while sharing different features. In such a case, the financial data are often vertically partitioned and owned by different financial institutions (e.g., a bank knows the user's account balance and an insurance company knows the user's lifestyle and habits). Since financial data are often represented as graphs, vertical federated GNNs are suitable for this scenario. 
Moreover, the features of users in the financial parties contain much private information, e.g., deposit amount, which highlights the importance of evaluating the robustness of vertical federated GNNs against privacy attacks, such as label inference attacks.


These are the premises of our work, whose objective is to design a novel advanced label inference attack against node-level classification using vertical federated GNNs. Compared to existing research, a fundamental characteristic of our attack is that it can work even in the worst setting for the attackers, i.e., when they have zero background knowledge about the possible relations between the labels and the reference dataset, which is why we named our solution BlindSage. The main limitation of existing label inference attacks~\cite{fu2022label} is that they typically require preliminary access to a percentage of pre-labeled data from the training set. For this reason, in many real-life scenarios, their applicability appears limited.
In addition to formalizing the attack definition, we conduct a comprehensive experimental analysis, demonstrating the efficacy of our attack design with various GNN architectures, datasets, and attackers' knowledge assumptions. Furthermore, we test potential mitigation strategies to enhance the robustness of VFL with GNNs against our proposal. We share our implementation at [link]\footnote{The link will be provided upon paper acceptance.}. 
In summary, the main contributions of this paper are:
\begin{itemize}
    \item We define a novel zero-background-knowledge attack, called BlindSage. Our approach exploits the characteristics of vertical federated GNNs on node classification tasks to craft an advanced and powerful attack strategy. Based on our experiments, BlindSage can achieve $100\%$ accuracy in most cases. In addition to such outstanding results, it is worth mentioning that, to the best of our knowledge, this paper is also the first work studying label inference attacks in vertical federated GNNs. 
    \item We relax the assumptions about the attackers' knowledge regarding the model's architecture and the number of classes, and we see that our attack can still reach a very high attack accuracy, i.e., more than $90\%$.
    \item We propose an early stopping strategy and provide an ablation study on different components of our attack, emphasizing their contribution.
    \item We test BlindSage in the presence of well-known defense methods. While some of these methods can mitigate the attack, they drastically compromise the performance of the main classification task. Therefore, we include an initial investigation of a possible strategy to design an effective defense. The preliminary results show the promising potential of our intuition.
\end{itemize}

This paper is organized as follows.
Section~\ref{sec:background} delves into the technical details about GNNs and vertical federated learning vital to understanding the proposed attack. 
Section~\ref{sec:methodology} presents our threat model and the label inference attack methodology. 
Section~\ref{sec:experiments} discusses the experimental setup and results of our attack experiments. 
Section~\ref{sec:defense} discusses potential defense mechanisms to counteract the label inference attack. 
Section~\ref{sec:related work} provides a comprehensive review of related works.
Finally, Section~\ref{sec:conclusion} concludes the paper and discusses possible future work.
Additional results for the number of attack iterations, early stopping strategy for the GAT architecture, and noise addition defense are provided in Appendices~\ref{sec: attack iterations analysis},~\ref{sec:early-stopping-gat}, and~\ref{sec:noise-defense}, respectively.

\section{Preliminaries}
\label{sec:background}

Here, we introduce the concepts needed for our attack.
In particular, we briefly describe GNNs and Federated Learning in general and then focus on its vertical implementation. 

\subsection{Graph Neural Networks (GNNs)}
\label{gnn}

Recently, GNNs have achieved significant success in processing non-Euclidean spatial data, common in many real-world scenarios~\cite{zhou2020graph}. Unlike traditional neural networks, e.g., Convolutional Neural Networks (CNNs) or Recurrent Neural Networks (RNNs), GNNs work on graph data. 
GNNs take a graph $G=(\boldsymbol{V}, \boldsymbol{E}, \boldsymbol{X})$ as an input, where $V, E, X$ denote all nodes, edges, and node attributes in the graph, and learn a representation vector (embedding) for each node $\boldsymbol{v} \in \boldsymbol{V}$, $z_{\boldsymbol{v}}$, or the entire graph, $z_G$. 

This work considers a node classification task, where the node representation, $z_{\boldsymbol{v}}$, is used for prediction. 
A node's representation is computed by recursive aggregation and transformation of feature representations of its neighbors. After $k$ iterations of aggregation, a node's representation captures structure and feature information within its $k$-hop network neighborhood~\cite{xu2018powerful}. Formally, the $k$-th layer of a GNN is described with Eqs.~\eqref{equ:gnn_agg} and~\eqref{equ:gnn_trans}, 

\begin{equation}
x_{\boldsymbol{v}}^{(k)} = AGG^{(k)}(\left \{ z_{\boldsymbol{v}}^{(k-1)}, \left \{z_{\boldsymbol{u}}^{(k-1)}|\boldsymbol{u}\in \mathcal{N}_{\boldsymbol{v}} \right\} \right \}),
    \label{equ:gnn_agg}
\end{equation}

\begin{equation}
    z_{\boldsymbol{v}}^{(k)} = TRANSFORMATION^{(k)}(x_{\boldsymbol{v}}^{(k)})
    \label{equ:gnn_trans},
\end{equation}
where $z_{\boldsymbol{v}}^{(k)}$ is the representation of node $\boldsymbol{v}$ computed in the $k$-th iteration. $\mathcal{N}_{\boldsymbol{v}}$ are 1-hop neighbors of $\boldsymbol{v}$, and $AGG(\cdot)$ is an aggregation function that can vary for different GNN models. $z_{\boldsymbol{v}}^{(0)}$ is initialized as node feature. The $TRANSFORMATION(\cdot)$ function consists of a learnable weight matrix and an activation function. In node classification, there are two training strategies: inductive and transductive. In inductive training, the nodes that need to be classified are not seen during training, whereas in the transductive setting, these nodes (but not their labels) are accessed during training. This work focuses on the transductive setting widely used in the node classification task~\cite{kipf2017semigcn, velickovic2018gat}. 
Moreover, we consider three representative models of the GNN family. In the following, we briefly describe these models and their differences.

\vspace{5pt}
\noindent
{\bf Graph Convolutional Networks (GCNs)}~\cite{kipf2017semigcn} are widely used for node classification. Let $d_{\boldsymbol{v}}$ denotes the degree of node $\boldsymbol{v}$. The aggregation operation in GCN is then given as:
\begin{equation*}
    x_{\boldsymbol{v}}^{(k)} \leftarrow \sum_{u \in \mathcal{N}_{\boldsymbol{v}}\bigcup {\boldsymbol{v}}} \frac{1}{\sqrt{d_{\boldsymbol{v}}d_{\boldsymbol{u}}}}z_{\boldsymbol{u}}^{(k-1)}.
\end{equation*}
GCNs perform a non-linear transformation over the aggregated features to compute the node representation at layer $k$: $z_{\boldsymbol{v}}^{(k)} \leftarrow ReLU(x_{\boldsymbol{v}}^{(k)}W^{(k)}).$
The original GCN algorithm is designed for semi-supervised learning in a transductive setting, and the exact algorithm requires that the complete graph Laplacian is known during training. 

\vspace{5pt}
\noindent
{\bf GraphSAGE}~\cite{hamilton2017inductiverepr_graphsage_reddit} algorithm was initially proposed as an extension
of the GCN framework to the inductive setting.
Specifically, there are three candidate aggregator functions in GraphSAGE, i.e., MEAN, LSTM, and Pooling aggregators. 
In our paper, we focus on the MEAN aggregator so that the overall aggregation function is:
\begin{equation*}
    x_{\boldsymbol{v}}^{(k)} = \textrm{MEAN}({z_{\boldsymbol{v}}^{(k-1)}}\cup{z_{\boldsymbol{u}}^{(k-1)}|\boldsymbol{u}\in \mathcal{N}_{\boldsymbol{v}}}).
\end{equation*}
In GraphSAGE, like GCN, a non-linear transformation $\sigma$ is applied over the aggregated features: $ z_{\boldsymbol{v}}^{(k)} = \sigma (W^{(k)}\cdot x_{\boldsymbol{v}}^{(k)}).$

In addition to the standard neighbor aggregation scheme mentioned above in Eqs.~\eqref{equ:gnn_agg} and~\eqref{equ:gnn_trans}, there exist other non-standard neighbor aggregation schemes, e.g., weighted average via attention in {\bf Graph Attention Network (GAT)}~\cite{velickovic2018gat}. 
Specifically, given a shared attention mechanism $a$, attention coefficients can be computed by: $ e_{\boldsymbol{v}\boldsymbol{u}} = a(Wz_{\boldsymbol{v}}^{(k-1)}, Wz_{\boldsymbol{u}}^{(k-1)}),$
which indicate the importance of node $\boldsymbol{u}$'s features to node $\boldsymbol{v}$. Then, the normalized coefficients can be computed by using the \textit{softmax} function: $\alpha _{\boldsymbol{v}\boldsymbol{u}} = softmax_{\boldsymbol{u}}(e_{\boldsymbol{v}\boldsymbol{u}}).$
Finally, the next-level feature representation of node $\boldsymbol{v}$ is:
\begin{equation*}
    z_{\boldsymbol{v}}^{(k)} = \sigma \left ( \frac{1}{P}\sum_{p=1}^{P}\sum_{\boldsymbol{u}\in\mathcal{N}_{\boldsymbol{v}}}\alpha _{\boldsymbol{v}\boldsymbol{u}}^pW^pz_{\boldsymbol{u}}^{(k-1)} \right ),
\end{equation*}
where $\alpha _{\boldsymbol{v}\boldsymbol{u}}^p$ are the normalized coefficients computed by the $p$-th attention mechanism $a^p$ and $W^p$ is the corresponding input linear transformation's weight matrix.

\subsection{Federated Learning}

Federated learning enables $C$ clients to train a global model $\model$ collaboratively without revealing local datasets. Unlike centralized learning, where a central server must collect local datasets before training, FL performs training by uploading the weights of local models ($\{\model^i \mid i \in C\}$) to a parametric server. 
Specifically, FL aims at optimizing a loss function:
\begin{equation*}
    \mathop {\min }\limits_\model \ \ell (\model ) = \sum\limits_{i = 1}^n \frac{{{k_i}}}{C}{L_i}( \model^i ),\ {L_i}(\model^i ) = \frac{1}{{{k_i}}}\sum\limits_{j \in {P_i}} {{\ell _j}(\model^i, x_j)},
\end{equation*}
where ${L_i}(\model^i)$ and $k_i$ are the loss function and local data size of $i$-th client, and $P_i$ refers to the set of data indices with size $k_i$.
At the $t$-th iteration, the training can be divided into three steps.
Initially, all clients download the global model $\model_t$ from the server.
After that, the \textit{local training} takes place, during which each client updates the global model by training with their datasets: $\model_t^i \leftarrow \model_t^i-\eta\frac{\partial L(\model_t, b)}{\partial \model_t^i} $, where $\eta$ and $b$ refer to learning rate and local batch, respectively. 
Finally, after the clients upload their local models $\{\model_t^i \mid i \in C\}$, the server updates the global model by \textit{aggregating} the local models. 

\subsection{Vertical Federated GNNs}
\label{VFGNNs}

\begin{figure}[!ht]
    \centering
    \includegraphics[width=0.32\textwidth]{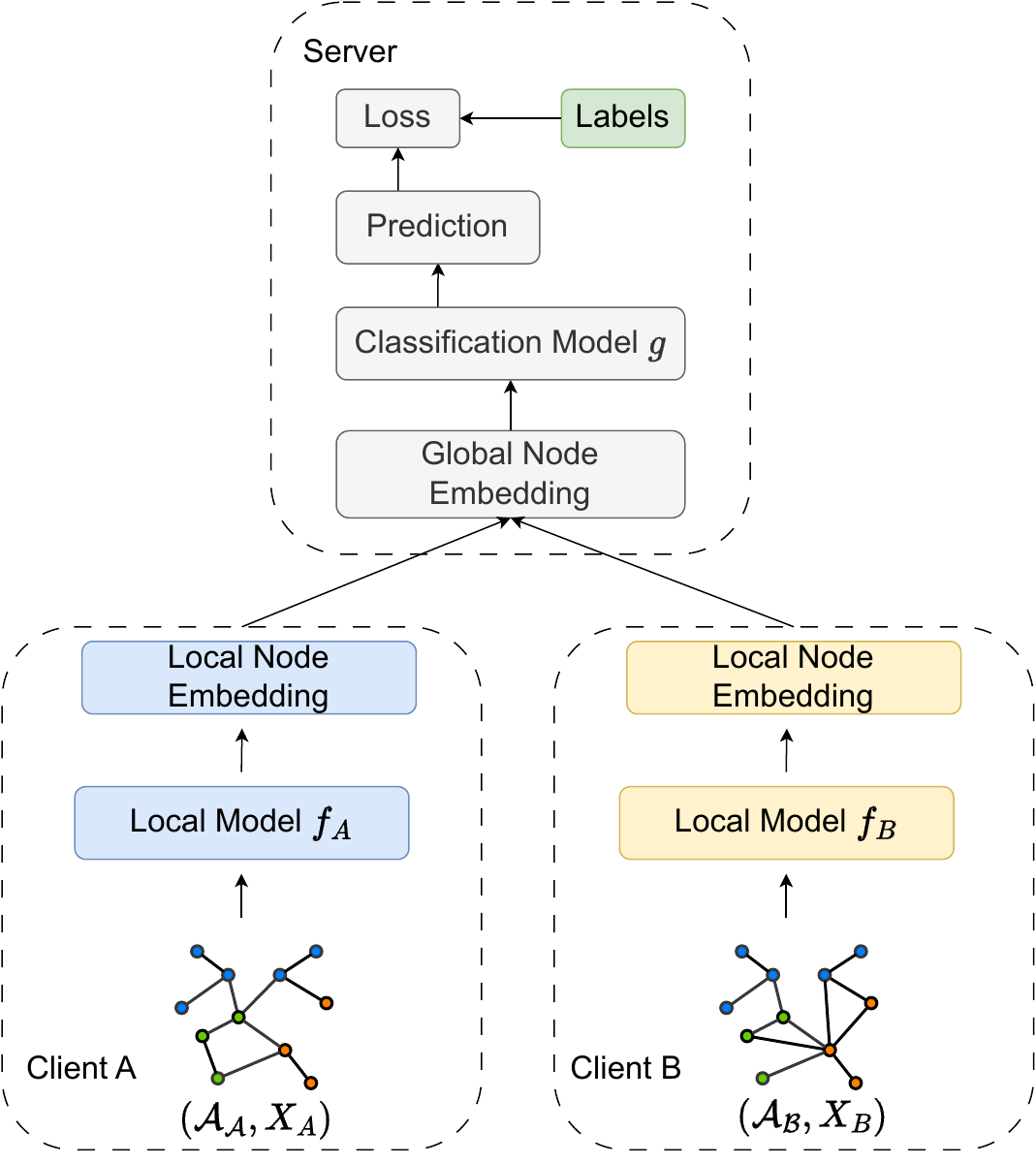}
    \caption{Illustration of vertical federated GNNs.}
    \label{fig:vertical_gnns}
\end{figure}

Vertical federated GNN is a federated GNN learning paradigm under vertically partitioned data. 
This paper focuses on vertical federated GNNs for the node classification task. 
A node-level vertical federated GNN consists of three components: local node embeddings, global node embeddings, and final predictions, as shown in Figure~\ref{fig:vertical_gnns}. 
A graph is split regarding its structure and node features, where each client knows partial node feature information and partial graph structure information. 
Each client generates local node embeddings through their local models. The local models use multi-hop neighborhood aggregation on graphs, as mentioned in Section~\ref{gnn}. For $\forall \boldsymbol{v} \in \boldsymbol{V}$ at each client, neighborhood aggregation is the same as the traditional GNN, i.e., $z_{\boldsymbol{v}}^{(k)}(i)$ is the output for all clients. 
The server that holds the label information combines the local node embeddings from clients and gets the global node embeddings $z_{\boldsymbol{v}}^{(k)}$. The combination strategy should maintain high representational capacity. 
One combination strategy is the \textit{mean} operation. It takes the element-wise mean of the vectors in ${z_{\boldsymbol{v}}^{(k)}(i), \forall i \in [1, C]}$ ($C$ is the number of clients), assuming clients contribute equally to the global node embeddings (Eq.~\eqref{eq:mean node embedding}).
\begin{equation}\label{eq:mean node embedding}
    z_{\boldsymbol{v}}^{(k)} \leftarrow MEAN({z_{\boldsymbol{v}}^{(k)}(i), \forall i \in [1, C]}).
\end{equation}

Another combination strategy is the \textit{concatenation} operation that fully preserves local node embeddings learned from different clients, as shown in Eq.~\eqref{eq:concat node embedding}. We chose this combination strategy for our experiments, but we believe our attack will also work if $MEAN$ is used. 
The comparison between the two strategies is provided in Appendix~\ref{sec:combinaton-aggregation}.

\begin{equation}\label{eq:concat node embedding}
    z_{\boldsymbol{v}}^{(k)} \leftarrow CONCAT(z_{\boldsymbol{v}}^{(k)}(1), z_{\boldsymbol{v}}^{(k)}(2), \cdots, z_{\boldsymbol{v}}^{(k)}(C)).
\end{equation}

The global node embeddings are then the input of a classification model $g$, which outputs the final predictions for the nodes. The labels held by the server are used to compute the loss and then update the classification model. The local models can be updated using the gradients they receive from the server. This VFL framework was also used in~\cite{chen2022graphfraudster}.

\section{Methodology}
\label{sec:methodology}

This section presents BlindSage and its threat model. We also explain our assumptions on the attacker's background knowledge and define our attack, including three strategies tailored to a particular attacker's knowledge level.

\subsection{Threat Model}
\label{sub:threatModel}

We assume that $C$ clients (where $C \geq 2$) jointly train a GNN model using the VFL framework described in Section~\ref{VFGNNs}. Each client holds partial node features and partial graph structure information. 
For example, in Figure~\ref{fig:vertical_gnns}, we can see $2$ clients, $A$ and $B$, and a graph $G$ with features $X$. The node feature information $X$ is split into two parts, $X_A$ and $X_B$, owned by client $A$ and $B$, respectively. 
Moreover, the graph structure information $\mathcal{A}$ is split into $\mathcal{A_A}$ and $\mathcal{A_B}$, owned by client $A$ and $B$, respectively. 
Client $A$ (resp. $B$) trains its local model $f_A$ (resp. $f_B$) based on the graph structure information $\mathcal{A_A}$ (resp. $\mathcal{A_B}$ ) and node feature information $X_A$ (resp. $X_B$). 
In particular, there is no overlap between the set of edges and the features owned by the clients. 
Moreover, each client controls the training and execution of its local model that generates the embeddings for the samples $H_A$ ($H_B$). The labels of the training dataset are privately owned by a server, which controls the training and execution of a classification model $g$. The loss $l$ is calculated in the server with the labels and then used to update the classification model $g$. 
The clients can act independently as \textit{adversary}, aiming to infer the private labels without affecting the federated model. In our scenario, one of the clients is the \textit{adversary}.

In VFL, as illustrated in Figure~\ref{fig:vertical_gnns}, at each training round, each client, say $A$, receives from the server the partial gradients $\nabla H_A$  of the loss $l$ concerning the embeddings generated by its local model.
Then, the client uses them to generate the gradients $\nabla W$ for its local model as follows ($CE=$ cross-entropy, $SM=$ \textit{softmax}):
\begin{gather}
    l = CE(SM(preds),\ SM(RealLabels)) \notag \\
    \nabla W =\sum \frac{\partial l}{\partial H_A} \cdot \frac{\partial H_A}{\partial f_A}. 
\end{gather}

Since the server calculates the gradients using the real labels, $RealLabels$, in the loss, we argue that the adversary can exploit the received gradients to conduct an attack during the training phase and extract useful information about them.
The approach may vary depending on the attacker's knowledge of the model architecture and the number of classes used for the classification.

In the Federated Learning field, especially in the context of Horizontal Federated Learning (HFL), two levels of knowledge are typically considered: {\em full} and {\em partial}.
In the first case, all the entities involved have access to information about the data distribution across the workers and the updates of each of them during the training.
In the second case, instead, the workers have knowledge solely of the data points they own and their local training updates, with only partial access to information about the system.
In our paper, we are targeting the latter scenario, assuming that only local information of the controlled nodes is available to the attacker. 
Moreover, in VFL, only the server trains the full classification model, while the clients participate by training a local model with no classification layer, so they do not have information about the classes and the full model architecture. Consequently, the knowledge of the learning objective (i.e., the classification model, including the output size), which is always known to all the clients in HFL, is, instead, critical information in VFL. 
Due to this characteristic of VFL, in practice, we need to extend the above taxonomy by adding three additional sub-levels of knowledge for the attacker, namely:
\emph{basic knowledge}, \emph{limited knowledge}, and \emph{no knowledge}.
In the first case, the attacker has basic knowledge of the learning objective, meaning that the attacker knows the number of classes $n$ and the exact architecture of the server model $g$. That is the most favorable case for the attacker.
In the second case, referred to as \textit{limited knowledge}, the attacker knows only the number of classes $n$ and has no information regarding the architecture of the classification model maintained by the server. 
In this case, the attacker can hypothesize a classification model, possibly with a different architecture than the server one. 
Then, the same strategy of the previous case can be adopted to estimate the association between the $n$ labels and the training data.
In this scenario, the attack's performance could be affected by the fact that the latent information about the labels in the loss estimated by the server may be difficult to extract because of the mismatch between the architecture of the original classification layer of the server and the one assumed by the attacker.
The last case represents the worst-case scenario for the attacker. In this case, they have neither information on the number of classes $n$ nor the architecture of the classification layer of $g$ (\textit{no knowledge}). In this setting, the attacker must approximate the server model as in the previous case and estimate the number of classes $n$ to perform the attack.
Our intuition is that node embeddings learned during the federated learning include crucial information about the classes the nodes belong to.
Intuitively, the embeddings of the nodes belonging to the same class must share more similar characteristics than those of other class nodes.
According to this reasoning, the number of classes can be obtained by adopting an unsupervised learning strategy (clustering algorithm) and identifying the optimal number of clusters for the node embeddings.
Once again, such a task can be carried out during the federated learning task but only after a sufficient number of learning epochs for the local model.
Indeed, for the above strategy to work, the embeddings must already contain sufficient information about the training data.
The following section will provide all the technical details of our attack strategy.

\subsection{BlindSage}
\label{sub:attack_framework}

As discussed in Section~\ref{sub:threatModel}, each client holds information about its local graph and related features. Clients cannot access the true labels of each node nor the data features of the other clients.
The attacker, controlling one of the clients, can leverage only the gradients sent by the server to each client $X$ to update the corresponding local models $f_X$, where $X$ represents any passive client (e.g., $X \in \{ A, B \}$).
As stated above, because the server has generated these gradients by calculating a cross-entropy (CE) loss over the predictions using the real labels associated with each node in the graph, they must intrinsically hold information about such labels. 
Thus, we first explain the attack concept and intuition, followed by a more formal description.

\subsubsection{Attack Concept}
\label{sec:attack-concept}
The intuition behind our label inference attack is to leverage the leakage from the gradients to steal the labels from the server. Using gradient leakage is a common strategy in federated learning settings, where such information is the only information shared between the parties involved in the system~\cite{zhu2019deep}. 
While in the beginning, gradients were considered safe to share, recently, researchers showed that this is not the case since they carry information about processed data along with the final outputs through the loss function~\cite{zhu2019deep}.
Our attack strategy is based on this concept; indeed, during the training of a federated model, the attack is carried out by leveraging only the gradients returned by the server without requiring any additional background information from other participants. 
For this reason, the attack can go unnoticed by utilizing only local information available on the controlled malicious client and the history of gradients collected during the training of the federated model.
\begin{figure*}[!ht]
    \centering
    \includegraphics[width=0.70\textwidth]{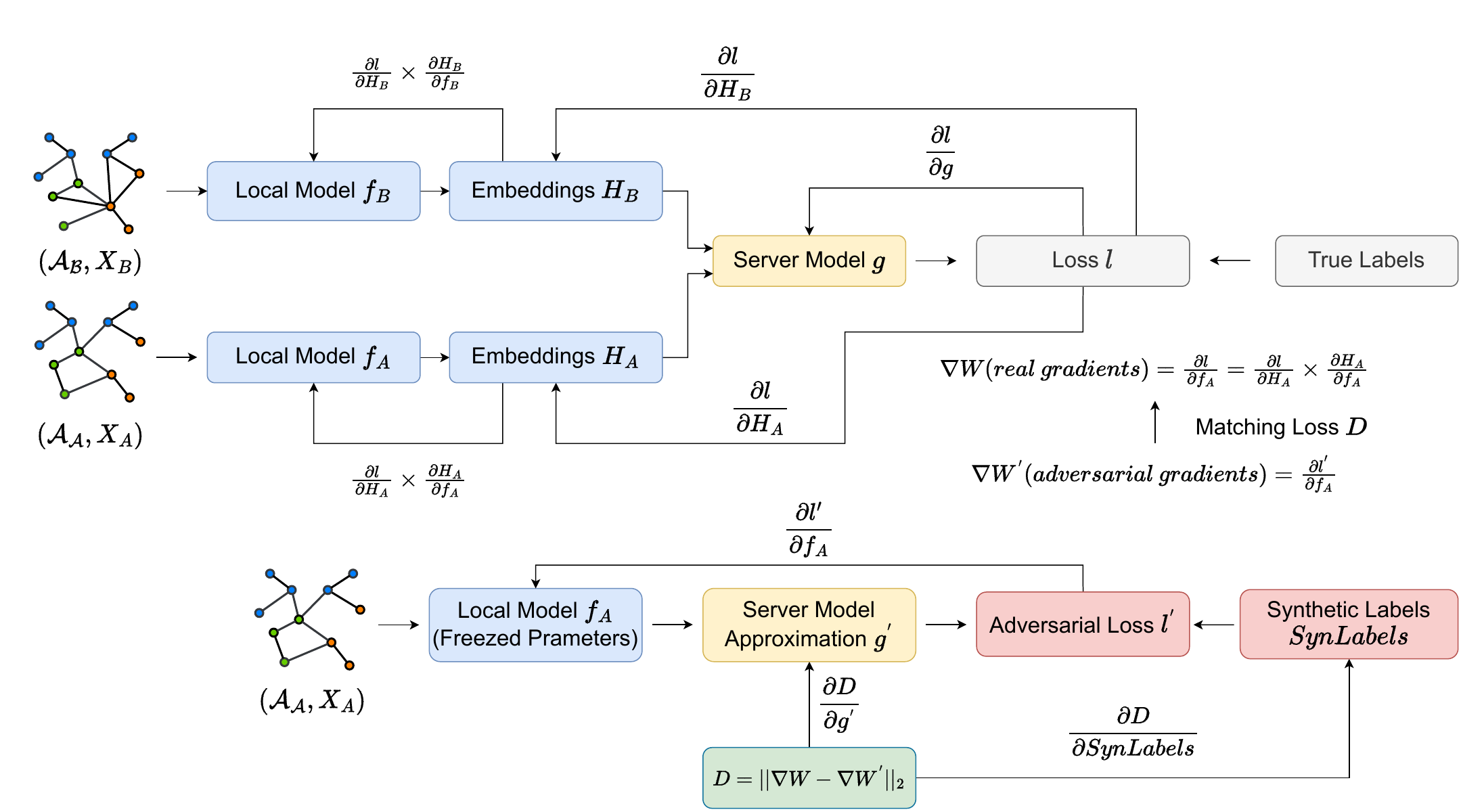}
    \caption{The BlindSage attack framework.}
    \label{fig:attack_framework}
\end{figure*}

In practice, as shown in Figure~\ref{fig:attack_framework}, assuming that the attacker controls a client $A$, the idea is to infer the private labels by trying to match the gradients $\nabla W$ for the local model $f_A$, computed based on the information returned by the server, with adversarial gradients generated by the attacker $\nabla W'$.
This strategy is similar to the one proposed in~\cite{zhu2019deep} used to recover the training data in HFL.
However, in HFL, an adversarial client controls a complete local model, including the gradients of all the parameters of the local model, while in VFL, the adversarial client controls only part of the federated model and has only access to the gradients of its local model which cannot run independently~\cite{fu2022label}, thus making the VFL setting more challenging.
In our scenario, to generate the adversarial gradients $\nabla W'$, the attacker must replicate the entire architecture, including an approximation for the server model $g'$ and the set of synthetic labels $SynLabels$.
$SynLabels$ are initialized as a tensor in which each element associated with a node of the input graph is an array with a length equal to the number of classes, and each element represents the probability for the node to belong to a specific class.
We initialize the tensor arrays so that each class has an equal probability, i.e., we set the starting value of each vector element to $1/n$.
According to the worst-case scenario presented in Section~\ref{sub:threatModel}, the number of classes $n$ can also be unknown to the attacker.
In this situation, we leverage an unsupervised learning algorithm to identify optimal clusters for the node embeddings produced by the local model $f_A$ controlled by the attacker. In our approach, we rely on the HDBSCAN algorithm~\cite{mcinnes2017hdbscan}. The rationale behind the choice of this algorithm is related to its characteristic of automatically estimating the optimal number of clusters, making it one of the best candidates for our purpose. However, we believe any clustering algorithm equipped with a heuristic for estimating the number of clusters can be exploited for this task (such as k-means with the silhouette method~\cite{dinh2019estimating}).
Depending on the scenario, the server model approximation $g'$ can be the exact clone of the classification network $g$ of the server in the \emph{basic knowledge} case or an approximation with a different architecture in the {\em limited knowledge} and {\em no knowledge} cases. 

\subsubsection{Formal Attack Description}

Without loss of generality, we formalize the attack with just one client, say $A$, controlled by the attacker, having access to only its local training data and local model $f_A$.
In the standard VFL approach, client $A$ generates the local node embedding $H_A$ using its local model $f_A$ and sends them to the server. Then, the server aggregates local node embeddings from all clients, including $H_A$, generating the final global node embedding $H$ (see Section~\ref{sec:background} for further details). 
The attack strategy, presented in Algorithm~\ref{alg:LIA}, is to replicate the training step at time $t$ as performed by the server, using only the local model $f_A$ with its parameters frozen at time $t-1$. 
In particular, the local node embeddings $H_A$ are given as input to the server model approximation $g'$, estimated by the attacker, which performs the predictions.
Exactly as done by the server when computing the global loss $l$, the above local predictions are then combined with the $SynLabels$ to calculate an adversarial loss $l'$; the adversarial loss is obtained using cross-entropy (CE) with a \textit{softmax} (SM).
The attacker uses $l'$ to generate the adversarial gradients $\nabla W'$ for the local model $f_A$ at time $t$. This is formulated in Eq.~\eqref{eqs:adversarial loss and gradient calculation}.
\begin{gather}
    preds=g'(local\_node\_embeddings) \notag\\	
    l'=CE(SM(preds), SM(SynLabel)) \label{eqs:adversarial loss and gradient calculation}, \nabla W' = \frac{\partial l'}{\partial f_A}. 
\end{gather}
From Eq.~\eqref{eqs:adversarial loss and gradient calculation}, observe that before passing $SynLables$ to the adversarial loss $l'$, we apply the \textit{softmax} function on both the local predictions and $SynLables$ to convert them into probabilities.
These gradients are, then, given as input to the matching loss\footnote{The function of matching loss is marked as \textit{ML} in Algorithm~\ref{alg:LIA}.} $D$ along with the real gradients $\nabla W$ at time $t$ (Eq.~\eqref{eqs:matching loss}).

\begin{equation}
\label{eqs:matching loss}
     D=|| \nabla W - \nabla W' ||_2.
\end{equation}

The obtained result is used to backpropagate the updates on the server model approximation $g'$ and to update the synthetic labels $SynLabels$, as follows:
\begin{gather}
    g' = g' - lr \cdot \frac{\partial D}{\partial g'}  \notag \\
    SynLabels = SynLabels - lr \cdot \frac{\partial D}{\partial SynLables}, 
\end{gather}

\noindent
where $lr$ is the learning rate of the optimizer used to update the $SynLabels$ and the parameter of $g'$.
This strategy aims at updating the server model approximation $g'$ and the $SynLabels$ to obtain the adversarial gradients $\nabla W'$ as close as possible to the real ones $\nabla W$.
The process above is repeated for a number of iterations $i$. Then, the obtained labels at the corresponding time $t+i$ are converted to a one-hot encoding to enhance the method's stability. Using a one-hot label encoding was initially proposed in the context of dataset distillation~\cite{sucholutsky2021soft}, where the authors showed that using such encoding at the beginning of each new iteration tends to increase the obtained accuracy. The improvement introduced by this strategy for our approach is presented in our ablation study in Section~\ref{sub:ablation}.

An essential observation regarding our solution is that the proposed attack operates stealthily without affecting other participating parties. 
That is achieved by exploiting only local information of controlled clients along with the gradients returned by the server according to the standard VFL mechanism. This characteristic allows the attacker to perform the attack in two modes: \emph{online} and \emph{offline}. In the former, the adversary carries out the attack during the execution of the global VFL task. In contrast, in the latter mode, the attacker can save the history of all the gradients produced by the training of the local model during the VFL task and subsequently execute the attack when the training ends. The \emph{offline} mode ensures no impact on the overall computational time of VFL, although it requires the storage of the gradient update history.

It is worth noting that our approach does not change across the different knowledge scenarios for the attacker. 
Indeed, it includes several pre-attack strategies that the attacker can use to leak additional knowledge about the scenario, using only local information.
In the following sections, we will provide insights on how the attacker, starting from a situation of \emph{no knowledge}, can perform pre-attacks to predict and estimate additional data, moving towards a \emph{basic knowledge} scenario. In particular, we will show how the attacker can {\em approximate} the architecture of the server model and {\em estimate} the number of classes. This strategy resembles model stealing and can be combined with our label inference approach. Using this additional information, the attacker can now perform the attack offline in \emph{basic knowledge} scenario. 
Thus, our attack can be executed by leveraging exclusively local information accessible within a single targeted node. Due to these characteristics, we term our approach a zero-background-knowledge attack, as it operates without needing any {\em prior} background information typically required by existing label inference attacks.

Our attack differs from the existing attacks~\cite{fu2022label} as it does not require a subset of labeled data and can be applied even when the attacker knows nothing about the server's architecture or the number of classes in the dataset. Our attack is also stealthy as it can be performed in an offline mode that does not affect the performance of the main VFL task. For that reason, we have designed a novel heuristic that identifies the stopping point of our algorithm in Section~\ref{sec:early_stopping}.

\begin{algorithm}[!ht] 
\scriptsize
\caption{BlindSage Label Inference Attack. \label{alg:LIA}}
\begin{algorithmic}[1]
\Require{\\ $\nabla W$ (Real Gradients) \\ $SynLables$ (Synthetic Labels) \\ $g'$ (Server Model Approximation) \\ $f_A$ (Local Model) \\ $X_A$ (Local Feature Data)} \\ $\mathcal{A_A}$ (Local Graph Structure)
\Statex
\While {stopping criterion not met}
    \For{$e$ in $AttackIterations$}   
        \State {$preds$ $\gets$ {$g'(f_A(\mathcal{A_A}, X_A))$}}
        \State {$l'$ $\gets$ $CE(SM(preds), SM(SynLabels))$}
        \State {$\nabla W'$ $\gets$ $l'.backward()$}
        \State {$D$ $\gets$ $ML(\nabla W, \nabla W')$}
        \State {$g'$ $\gets$ {$D.backward()$}}
        \State {$SynLabels$ $\gets$ {$D.backward()$}}
    \EndFor
    \State {$SynLabels$ $\gets$ {$OneHotEncoding(SynLabels)$}}
\EndWhile
\end{algorithmic}
\end{algorithm}

\section{Evaluation of the Proposed Attack}
\label{sec:experiments}

This section presents the experiments and the analysis we conducted to validate BlindSage's effectiveness. We measure BlindSage's performance under different GNN architectures and levels of knowledge of the attacker. We also propose two techniques to estimate the best architecture of the adversarial local model $g'$ adopted by the attacker to clone the VFL server model. Finally, we perform an ablation study on the different components of our attack.

\subsection{Experimental Setup}
\label{sub:experimental_setup}

In our experiments, we focused on a testbed in which: {\em (i)} the main VFL task is node classification; {\em (ii)} the server building the global model is the only party having complete information about the actual labels for the nodes; and {\em (iii)} the involved clients train local models on their private data, which are related to the same samples but include different features. 
All the experiments have been performed by training the global network of the VFL task with a learning rate of $0.01$, as done in~\cite{chen2022graphfraudster} using three architectures described in Appendix~\ref{sec:architectures}, in particular GCN, GAT and GraphSage.
We tested our approach on five different datasets, described in detail in Appendix~\ref{sec:datasets}, namely Cora, Citesser, Pubmed, Polblogs, and Reddit.
The attack, instead, has been tested using three different values of the learning rate, namely, $\{ 0.1, 0.5, 1 \}$. This allowed us to obtain very satisfactory performance under different assumptions on the attacker's knowledge. Indeed, based on our results, a higher learning rate was needed in scenarios where limited knowledge is assumed. Intuitively, this is needed to magnify the signal of larger graphs, as we discussed in Section~\ref{sec:early_stopping}. 
In the following, we report the results of the best-performing learning rate according to the considered dataset, architecture, and scenario. 
The matching loss used to calculate the distance between the real gradients $\nabla W$ and adversarial gradients $\nabla W'$ is an $L2$ norm (see Eq.~\eqref{eqs:matching loss}).
To evaluate the performance of our approach, we used the top-1 accuracy in the scenarios in which the considered dataset is Cora, Citesser, Pubmed, and Polblogs. For the Reddit dataset, we also used the top-3 and top-5 accuracy to validate our attack, as commonly done in the related literature for more complex datasets~\cite{fu2022label}.
Since the attack aims to extract the labels, the accuracy of a label inference attack is measured as the percentage of labels the attacker can correctly infer. According to this metric, if a label inference attack has an accuracy of 80\%, the attacker can correctly infer the labels of 80\% of the data used to train the model. 
Top-$k$ accuracy considers the attack successful if the correct class appears within the top $k$ predictions based on the \textit{softmax} output probabilities.
In all the scenarios, the number of iterations for the attack for each epoch during the VFL task, namely $AttackIterations$ in the Algorithm~\ref{alg:LIA}, is set equal to $10$. This setting is based on our initial experiments reported in Appendix~\ref{sec: attack iterations analysis}. 
The architecture of the server model $g$, in the {\em basic knowledge} case, matches the local model of the attacker $g'$. 
In the cases of {\em limited knowledge} and {\em no knowledge}, for the sake of simplicity, we defined $g'$ by modifying the original model $g$ by adding one more layer.
However, in Section~\ref{sec:server-model-approximation}, we thoroughly analyze the approximation of the server model $g'$ with a pool of four different classification models. 

\subsection{Performance Evaluation of the BlindSage Attack}
\label{sec:results}

This section presents BlindSage's results against the three different local model architectures presented in Appendix~\ref{sec:architectures} and the datasets presented in Appendix~\ref{sec:datasets}.
Moreover, once again, we considered three scenarios characterized by a different knowledge level of the attacker, namely: {\em basic knowledge}, {\em limited knowledge}, and {\em no knowledge}.
In Table~\ref{tab:results}, we report the obtained accuracy of our attack. 

As expected, in the case with {\em basic knowledge}, our attack achieved the best performance, with an accuracy equal to or close to $100\%$ in all cases.
Even in the scenarios with {\em less} or {\em no knowledge}, our attack archived very satisfactory results, obtaining over a $90\%$ accuracy in most cases.
As discussed in Section~\ref{sec:attack-concept}, HDBSCAN approximates well the number of classes in the dataset, leading to a good attack performance even when the attacker has no information on the number of labels. Furthermore, as explained above, in this experiment, $g'$ differed from $g$ in only one level. In Section~\ref{sec:server-model-approximation}, we will assess how the server model $g$ can be approximated by testing different architectures of $g'$ and choosing the one that obtains the best attack performance.

In the {\em no knowledge} case, performance is expected to be lower due to the requirement of having partially trained local node embeddings before accurate clustering is possible (see Section~\ref{sec:attack-concept}). The attacker must delay the exploit until a few epochs of the main task have passed for local embeddings to be partially learned. 
Section~\ref{sec:early_stopping} will show that the initial VFL epochs are most crucial for the attack as they yield the highest gradients. Thus, for a successful exploit, the attacker must determine the minimum epochs necessary for the local model to train before initiating the attack.

\begin{table}[!ht]
\centering
\caption{Accuracy of the proposed attack using different GNN architectures on the considered datasets.}
\label{tab:results}
\resizebox{\columnwidth}{!}{%
\begin{tabular}{ccccc}
\toprule
         &   & Basic Knowledge & Limited Knowledge & No Knowledge \\
Architecture         & Dataset  & lr=0.1 & lr=0.5 & lr=1 \\
\midrule
\multirow{4}{*}{GCN} & CORA     & 100\%          & 100\%               & 93\%          \\
                     & Citeseer & 100\%          & 100\% (lr=1)              & 87\%          \\
                     & Pubmed   & 100\% (lr=0.5)          & 98\%               & 96\%          \\
                     & Polblogs & 100\% (lr=0.5)          & 99\%               & 95\%          \\
\midrule
\multirow{4}{*}{GAT} & CORA     & 100\%           & 98\%               & 95\%          \\
                     & Citeseer & 100\%     & 90\%               & 88\%          \\
                     & Pubmed   & 100\% (lr=1)         & 96\% (lr=1)              & 93\%          \\
                     & Polblogs & 100\% (lr=0.5)       & 98\%               & 90\%          \\
\midrule
\multirow{4}{*}{GraphSage} & CORA     & 100\%     & 100\%               & 98\%          \\
                     & Citeseer & 100\%          & 100\%               & 98\%          \\
                     & Pubmed   & 90\% (lr=1)         & 90\% (lr=1)              & 86\%          \\
                     & Polblogs & 100\% (lr=0.5)         & 99\%               & 99\%          \\
\bottomrule
\end{tabular}%
}
\end{table}

Finally, using only GCN as a reference architecture, we tested BlindSage on a much larger training set, i.e., the Reddit and Arxiv datasets introduced in Appendix~\ref{sec:datasets}. The reason for choosing GCN as reference architecture is, once again, to keep the experiment feasible from a computational point of view, considering the large dataset size and the hardware setting of the machine adopted for our tests (see Section~\ref{sub:experimental_setup}).
The GAT and GraphSage architectures necessitate storing a dense graph representation, demanding 404 GB of RAM in our case with Reddit, and it was unfeasible to load such data on our experimental hardware.
The obtained results are reported in Table~\ref{tab:resultsReddit}. 
Our attack, even for a much larger dataset, still achieved a top-5 accuracy of around $70\%$. The top-k accuracy metric is typically used by classification approaches to assess the performance in the case of datasets with a large number of labels due to the high complexity of the problem~\cite{fu2022label,zou2022defending}. We also see that BlindSage's behavior is similar for all top-$k$ accuracy metrics. Its performance reaches the best values in the \emph{basic knowledge} scenario.
As expected, the absolute accuracy is improved as we increase the considered top $k$ predictions for accuracy.
Despite accuracy being lower than $100\%$, thus making the attack not fully effective, the number of correctly retrieved labels allows for subsequent passive label inference attacks~\cite{fu2022label}. In a passive attack, the adversary needs a subset of the true labels to mount a stronger label inference attack. Hence, combining BlindSage with the attack of~\cite{fu2022label} would allow the attacker to carry out a complete and effective label inference attack successfully. 

\begin{table}[!ht]
\centering
\caption{Accuracy of our attack on the Reddit dataset (the learning rate was $1$ for all the obtained results).}
\label{tab:resultsReddit}
\resizebox{\columnwidth}{!}{%
\begin{tabular}{cccccc}
\toprule
Architecture        &  Dataset                   & Attacker Knowledge & Top-1 Acc & Top-3 Acc & Top-5 Acc \\
\midrule
\multirow{3}{*}{GCN}&      \multirow{3}{*}{Reddit}     & Basic Knowledge & 32\% & 61\% & 72\%       \\
                    &                                & Limited Knowledge                  & 30\% & 56\% & 70\%       \\
                    &                               & No Knowledge                       & 28\% & 55\% & 68\%       \\
\midrule
\multirow{3}{*}{GCN}&      \multirow{3}{*}{Arxiv}     & Basic Knowledge                 & 37\% & 63\% & 76\%       \\
                    &                                & Limited Knowledge                  & 36\% & 61\% & 74\%       \\
                    &                               & No Knowledge                       & 35\% & 60\% & 73\%       \\
\bottomrule
\end{tabular}%
}
\end{table}

\subsection{Early Stopping Strategy for BlindSage}
\label{sec:early_stopping}

During the attack itself, the attacker is unaware of its performance. Understanding whether it should be carried out for the entire VFL task or stopped earlier could ensure better attack performance.
To make such a decision, the attacker has only access to the gradients returned by the server.
Therefore, we analyzed the attack accuracy against the local model average gradient values at each training step to identify a suitable early stopping strategy.
In the first row of Figures~\ref{fig:grad_accuracy_gcn},~\ref{fig:grad_accuracy_gs}, and~\ref{fig:grad_accuracy_gat} in Appendix~\ref{sec:early-stopping-gat}, we report the attack accuracy using different learning rates. We used solid lines to indicate the best-performing learning rate in these figures.
We plot the average value of the local model gradients in the second row of such images.
\begin{figure*}[!ht]
    \centering
    \includegraphics[width=0.76\textwidth]{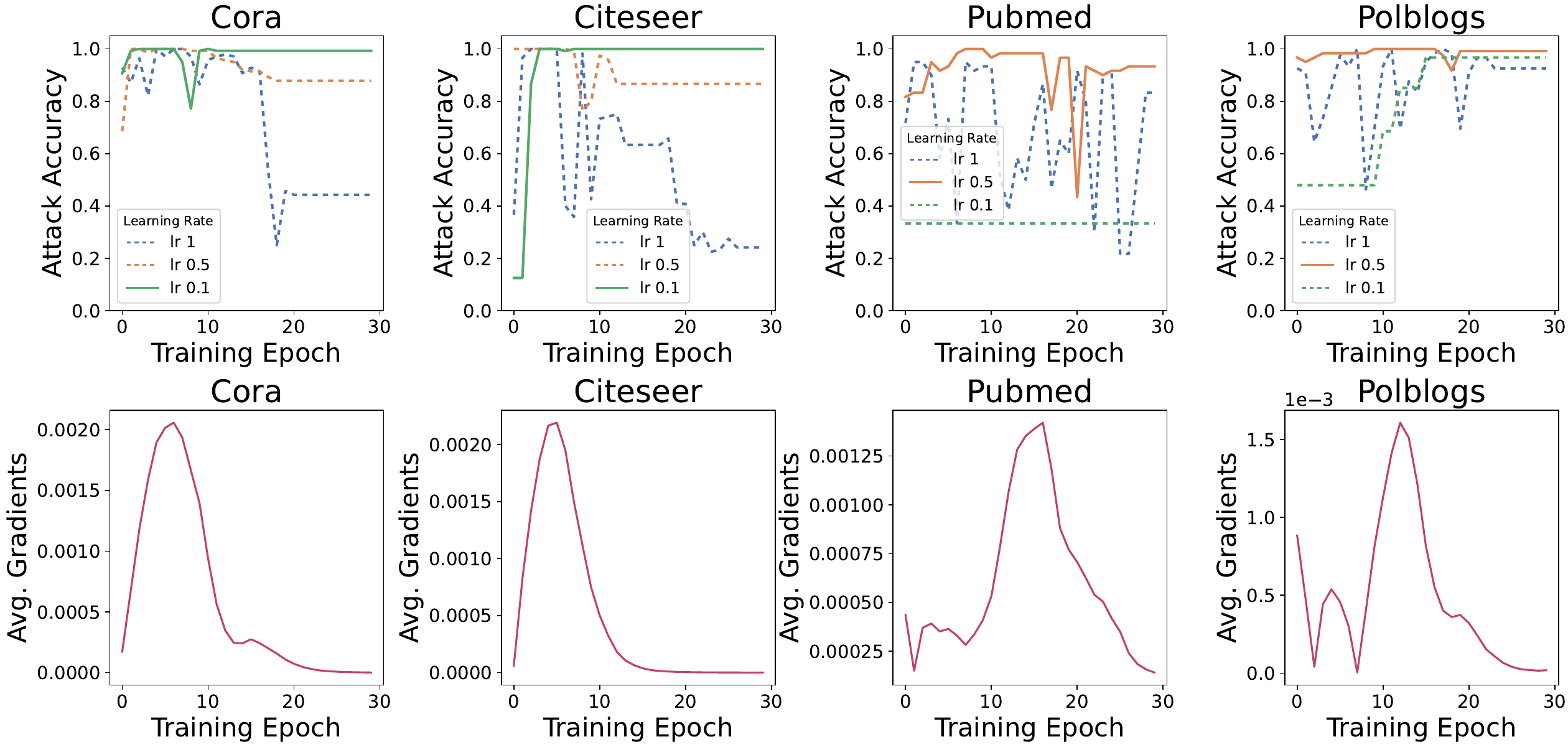}
    \caption{Attack accuracy versus the average values of the local model gradients in a {\em basic knowledge} scenario using the GCN architecture.}
    \label{fig:grad_accuracy_gcn}
\end{figure*}


\begin{figure*}[!ht]
    \centering
    \includegraphics[width=0.76\textwidth]{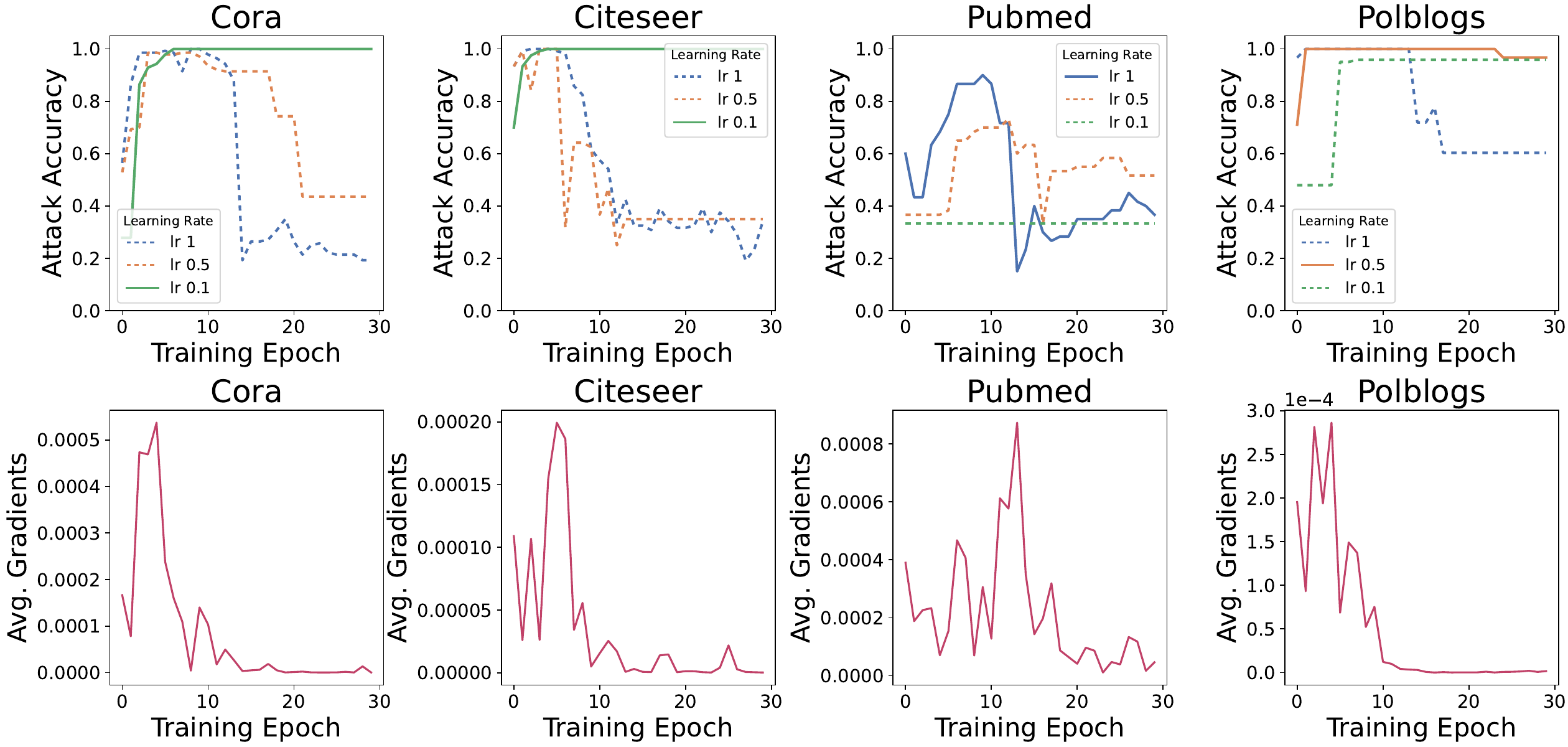}
    \caption{Attack accuracy versus the average values of the local model gradients in a {\em basic knowledge} scenario using the GraphSage architecture.}
    \label{fig:grad_accuracy_gs}
\end{figure*}
This experiment is independent of the scenario considered. Indeed, because gradients relate solely to the training of the VFL model, our investigation is not influenced by the particular attacker knowledge level considered. As a result, any knowledge configuration of the attacker should yield consistent patterns. Therefore, we opted to consider only the {\em basic knowledge} scenario for this experiment, simplifying the analysis.
From the obtained results, we can clearly see a correspondence between a spike in the magnitude of the gradients and a significant decline in the accuracy of the attack. Since our attack is highly dependent on the sign of the gradients, when the gradients' values start to decrease, it impacts our attack, recording a negative spike in the performance.
Usually, a few epochs later than the spike, the accuracy of the attack returns to a higher level because it adapts to the sign of the gradients, even if the average value of the gradients keeps decreasing. As we will see in Section~\ref{sec:defense} and Appendix~\ref{sec:noise-defense}, this result can be used to define a mitigation strategy for our attack based on a frequent variation of the gradients.
However, most of the time, the accuracy only partially recovers to the same level observed before the spike in the gradient magnitude. 
It is important to observe that the attack is highly dependent on the optimization process, the sign, and the
magnitude of the produced gradients. When the gradients changed their sign, we recorded a systematic drop in the performance of the attack. This is even more evident for configurations with higher learning rates because, in these cases, the gradients tend to fluctuate more when the loss function reaches the possible minimum value, thus causing a detriment in the informative content of the gradients themselves for the attacker.
Another important observation is that, particularly in settings with high learning rates, a noticeable performance decline is experienced when gradients become excessively small. This decline can be attributed to the matching loss computed on gradients, which, in this case, lacks attack-relevant information. This finding further underscores the rationale for implementing an early-stopping policy.
Previously, in this section, we saw that the maximum value of the gradients typically leads to the best performance for the attack for all the learning rates tested. 
Therefore, the attacker can monitor the gradients' trend and maintain the attack results obtained a few steps before the maximum average gradient value.
Another important point that we noticed is that once the gradients become too small, e.g., close to the end of the training, the accuracy of the attack is stable or slightly decreases; this happens mainly with higher learning rates.
This result confirms our intuition that the most critical epochs are the early ones in which the model learns the most, thus providing the most informative gradients to carry out our attack.
Another aspect to discuss is the performance of our solution on Pubmed, PolBlogs, and Reddit datasets with a learning rate of 0.1. The results show notable reductions in performance when the underlying graph of the GNN grows. This is due to the challenge of compressing information in a small space for larger graphs, resulting in a weakened exploitable signal from gradients. A low learning rate amplifies this issue, making satisfactory performance achievable only with medium/high learning rates (e.g., greater than 0.5) in these configurations.

\subsection{Server Model Approximation}
\label{sec:server-model-approximation}

The experiment described in this section focuses on assessing the performance of our attack against the different architectures for the local model. As presented in Section~\ref{sub:attack_framework}, the attacker must approximate a server clone model $g'$ to perform the attack. So far, in the previous experiment, we considered a static setting in which $g'$ differs from $g$ for only a layer ({\em limited} and {\em no knowledge}). 
However, as stated above, we expect that more complex architectures of $g'$ compared to $g$ would lead to a non-negligible impact on the attack performance.

\begin{figure}[!ht]
    \centering
    \includegraphics[width=0.74\columnwidth]{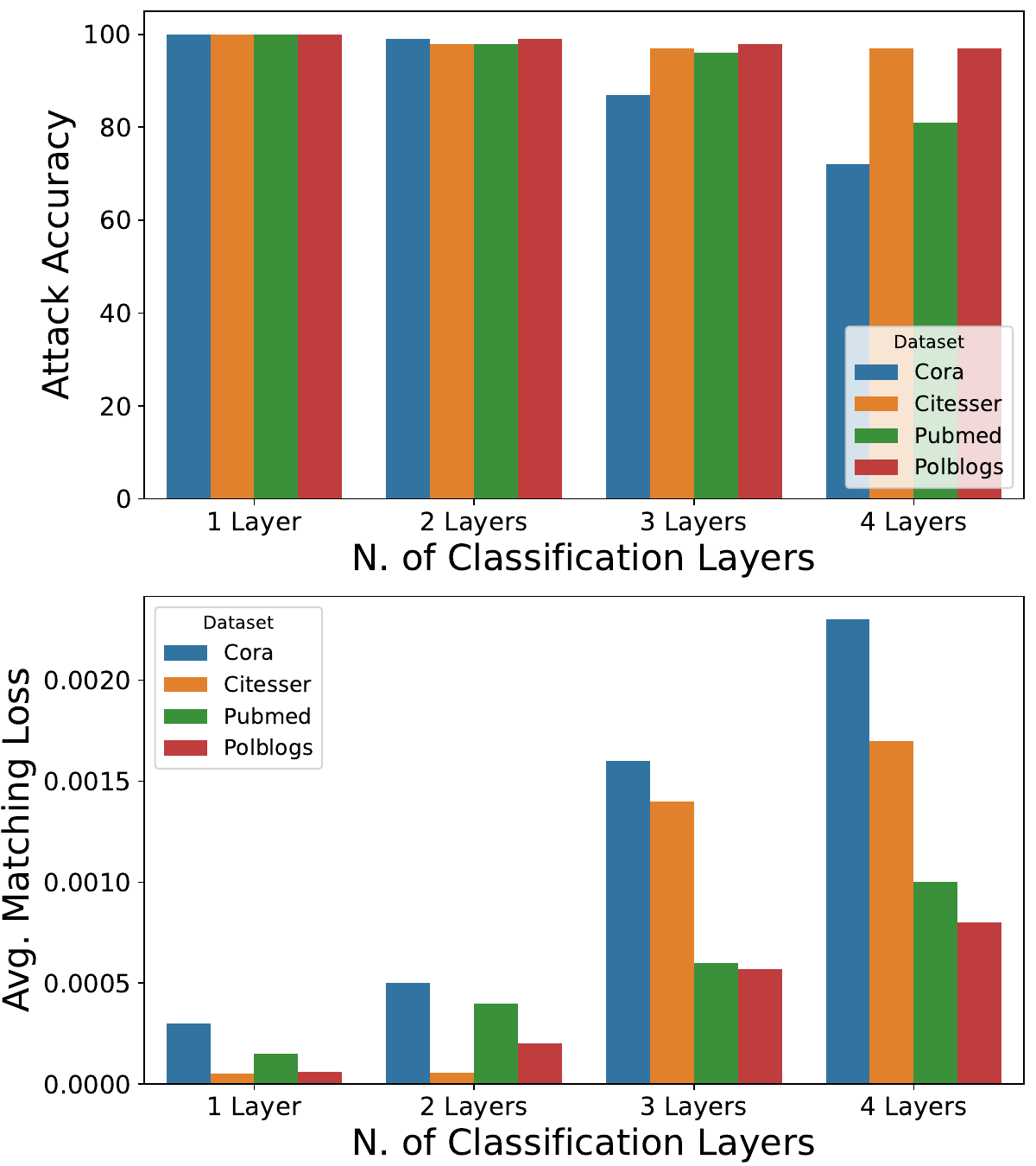}
    \caption{Comparison of the attack accuracy with the matching loss for the different $g'$ architecture on the considered datasets.}
    \label{fig:matching_loss}
\end{figure}

Therefore, this experiment analyzes different techniques an attacker can follow to identify the best server model clone $g'$.
To do so, first, we analyzed how the matching loss changes according to the number of layers of the server model approximation $g'$. In particular, using a classifier of one layer in the server, we ran our attack assuming a different number of layers for $g'$.
This exploration aimed to confirm our intuition that using within a set of candidate architectures the one with the lowest matching loss on the adversarial gradients $\nabla W'$ during the attack will lead to better attack performance.
Hence, we selected four different architectures for $g'$: one with a single classification layer like the real $g$ model as reported in Section~\ref{sub:experimental_setup}, and the others with two, three, and four layers, respectively. The rationale underlying the maximum number of layers for $g'$ comes from the observation that the dimension of a typical classifier $g$ is usually limited to a few layers~\cite{chen2022graphfraudster,fu2022label, chen2020vertically}. 
In Figure~\ref{fig:matching_loss}, we reported the results of this analysis conducted using the GCN architecture as a reference for the local model and the following datasets on which the attack performed the best to conduct the most objective analysis: Cora, Citeseer, Pubmed, and Polblogs.
As expected, the best performance is obtained for the architecture of $g'$, which has the lowest average matching loss.
Additionally, the architecture with the lowest average matching loss is the actual architecture of model $g$. 
As further confirmation, we tested our approach in the opposite scenario using a classifier $g$ of four layers. In particular, we evaluated our approach considering the Cora and Citeseer datasets. For both, we obtained $100\%$ of accuracy by selecting for $g'$ the architecture with the lowest matching loss in the set, respectively, a classifier with one layer for the former and a two-layer classifier for the latter.
This finding shows that our attack can be successful even if $g'$ differs from $g$ as the matching loss is more crucial than the actual server model architecture. It is worth noting that the matching loss uses only the gradients returned from the server; thus, it can be used in the \emph{no knowledge} scenario. From this experiment, we also conclude that even in the \emph{basic knowledge} scenario, the user should check the matching loss of a few server model architecture variations even if $g$ is known, as this could lead to a better attack.

The results described above provide new insights that can be used to improve our attack.
First, we can consider using different candidate architectures for $g'$ and switch between them during the VFL training to identify the one that obtains the lowest matching loss on the generated adversarial gradients $\nabla W'$. Observe that, to avoid impacts on the VFL performance, the above strategy can be performed offline. The only requirement for the attacker is to save all the intermediate gradients during the VFL training so that such information can be used later to infer the best matching architecture for $g'$.
The attacker can leverage an additional technique that uses an ensemble of the different candidate architectures for $g'$ to carry out an even more refined attack.
To do so, the attacker selects a pool of candidate architectures for $g'$ and, with each one of them, performs the predictions on the embeddings of the node. Then, all the produced predictions contribute to calculating the loss to generate the adversarial gradients $\nabla W'$.

We compared the two techniques above against the default strategy in which a static approximation of the architecture of $g'$ is guessed. For the lowest matching loss method, we leveraged the online mode. We tested these methods in a scenario {\em limited knowledge} of the attacker (to focus only on the classifier model identification), and we considered all three GNN architectures and the four datasets.
The results are reported in Table~\ref{tab:matching_loss_ensamble}.
As we can see, the two proposed strategies, in most cases, improved the performance of the default static strategy, reaching even 100\% attack accuracy.
Only for Pubmed, we see that the two proposed strategies may sometimes not lead to perfect results. 
We believe this is due to the large size of the Pubmed dataset, which negatively affects the attack performance.
In summary, the results obtained in this experiment allowed us to introduce two important strategies that are not strictly necessary to perform the attack but that the attacker can leverage to select the most appropriate model $g'$ to improve the performance of the BlindSage.

\begin{table}[!ht]
\tiny
\centering
\caption{Attack accuracy changing the technique adopted to approximate the model $g$.}
\label{tab:matching_loss_ensamble}
\resizebox{\columnwidth}{!}{%
\begin{tabular}{ccccc}
\toprule
Architecture         & Dataset  & Static & Lowest Matching Loss & Ensemble \\
\midrule
\multirow{4}{*}{GCN} & CORA     & 100\%            & 100\%               & 100\%          \\
                     & Citeseer & 100\%            & 100\%               & 100\%          \\
                     & Pubmed   & 98\%            & 100\%               & 98\%          \\
                     & Polblogs & 99\%            & 100\%               & 100\%          \\
\midrule
\multirow{4}{*}{GAT} & CORA     & 98\%            & 100\%               & 100\%          \\
                     & Citeseer & 90\%            & 100\%               & 100\%          \\
                     & Pubmed   & 96\%            & 100\%               & 100\%          \\
                     & Polblogs & 98\%            & 100\%               & 100\%          \\
\midrule
\multirow{4}{*}{GraphSage} & CORA     & 100\%     & 100\%               & 100\%          \\
                     & Citeseer       & 100\%     & 100\%               & 100\%          \\
                     & Pubmed         & 90\%     & 90\%                & 92\%          \\
                     & Polblogs       & 99\%     & 100\%               & 100\%          \\
\bottomrule
\end{tabular}%
}
\end{table}

\subsection{Ablation Study on Attack's Components}
\label{sub:ablation}

In this section, we present an ablation study to show the importance of each component that composes our solution.
We conducted the ablation study using the GCN architecture in a scenario where the attacker has {\em basic knowledge}.
To assess BlindSage's performance, we considered three different configurations with results reported in Table~\ref{tab:ablation}.
In the first one, we disabled the training of the server model approximation $g'$ by freezing the randomly initialized parameters. Correctly approximating the server model is, of course, essential since the attack matches the gradients to be able to extract the label information. However, this is also a very expensive task for the attack. Thus, we assess the importance of training this approximation model in our ablation study. 
As we can see from Table~\ref{tab:ablation}, the training of $g'$ using the matching loss plays a crucial contribution to the performance of the complete solution. The accuracy dropped to around 50\% when using the solution in which the training of $g'$ is frozen.

In the second configuration, we removed the application of the \textit{softmax} function to test its contribution to the attack performance. From the results, we can see that since the loss $l$ presented in Section~\ref{sub:threatModel} is calculated by applying the \textit{softmax} activation on the predictions, using only the one-hot encoding is not enough to obtain similar adversarial gradients $\nabla W'$ to match $\nabla W$. This causes a non-negligible detriment in the attack performance, thus confirming the crucial role of the \textit{softmax} function in our attack design.

As stated in Section~\ref{sec:attack-concept}, similarly to what was done in~\cite{sucholutsky2021soft}, we applied a one-hot encoding on the label after each iteration to improve the attack performance. 
The remaining configuration verifies the contribution of the one-hot encoding of the $SynLabels$ after the final loop of a given epoch.
The results also show the advantage given by the one-hot encoding before performing the attack on the final performance. 

\begin{table}[!ht]
\scriptsize
\centering
\caption{Ablation study of several BlindSage's components.}
\label{tab:ablation}
\begin{tabular}{ccccc}
\toprule
Scenario &                CORA &    Citeseer             & Pubmed & Polblogs \\
\midrule
No $g'$ model traning         & 46\%     & 38\%                & 56\%   & 56\%     \\
No \textit{softmax} function   & 85\%     & 99\%                & 93\%   & 98\%     \\
No one-hot encoding     & 91\%     & 95\%                & 80\%   & 100\%    \\
Full solution           & 100\%     & 100\%               & 100\%   & 100\%      \\  
\bottomrule
\end{tabular}
\end{table}

From the above results, we can state that the contribution of all the analyzed components is essential to achieve an effective attack.  
Among all, the correct approximation of the model $g'$ proved to be the most influential factor for the attack's success. Then, we repeated the test across the considered scenarios, confirming consistent behavior.

\section{Potential Defenses Against BlindSage}
\label{sec:defense}

This section assesses BlindSage's effectiveness against privacy-preserving mechanisms that the server can employ to prevent information leakage from the gradients.
One of the most common strategies to prevent such information leakage relies on applying Differential Privacy (DF)~\cite{liu2022federated, qi2020privacy}.
Differential privacy tries to preserve privacy by clipping the gradients' contribution to a specific threshold and adding Gaussian or Laplacian noise to obfuscate the information the attacker can infer.
Changing the clipping threshold and the level of noise added is a trade-off between performance and privacy. 
In the following experiment, we considered the two techniques separately. 
Limiting the information held by the gradients could prevent the attacker from inferring the labels, but at the same time, it could also affect the performance of the federated model.
Similarly to the work in~\cite{fu2022label}, another possible defense we tested against our attack is Gradient Compression (GC)~\cite{kairouz2021advances}.
The idea of gradient compression is to hide some of the gradients from the passive parties.
In particular, only the gradients with values larger than a given threshold are shared with the parties.
The idea is to preserve privacy by providing the passive party with only the minimum amount of information required to train the model, assuming it is enough to prevent attacks like the one we propose.
Another defense that acts on the gradients is the Privacy-Preseving Deep Learning (PPDL)~\cite{shokri2015privacy}. This defense adds noise to a random gradient value and sets to zero the gradient values smaller than a threshold. The process is repeated until a fraction $\theta$ of gradient values is altered. An additional defense to our attack can be DiscreteSGD~\cite{bernstein2018signsgd}. The server analyzes the mean and standard deviation of the distribution of the gradient to identify a ``safe'' interval and then discards the outliers. After that, the interval is sliced into N sub-intervals so that the gradients are rounded to the closest boundary of the sub-intervals before sending them to participants. The last considered defense is Label Differential Privacy (LabelDP)~\cite{ghazi2021deep}. In this case, the protection strategy introduces noisy labels to add uncertainty to the model instead of working directly on the gradients. The idea is to redistribute the probability of the correct labels across the other available categories, still guaranteeing a higher probability to the correct one, in such a way as to introduce a controlled uncertainty level on the classification. This uncertainty is, hence, propagated to the gradients, sent back to the malicious client, to inhibit inference attacks.

\begin{figure}[!ht]
    \centering
    \includegraphics[width=0.98\columnwidth]{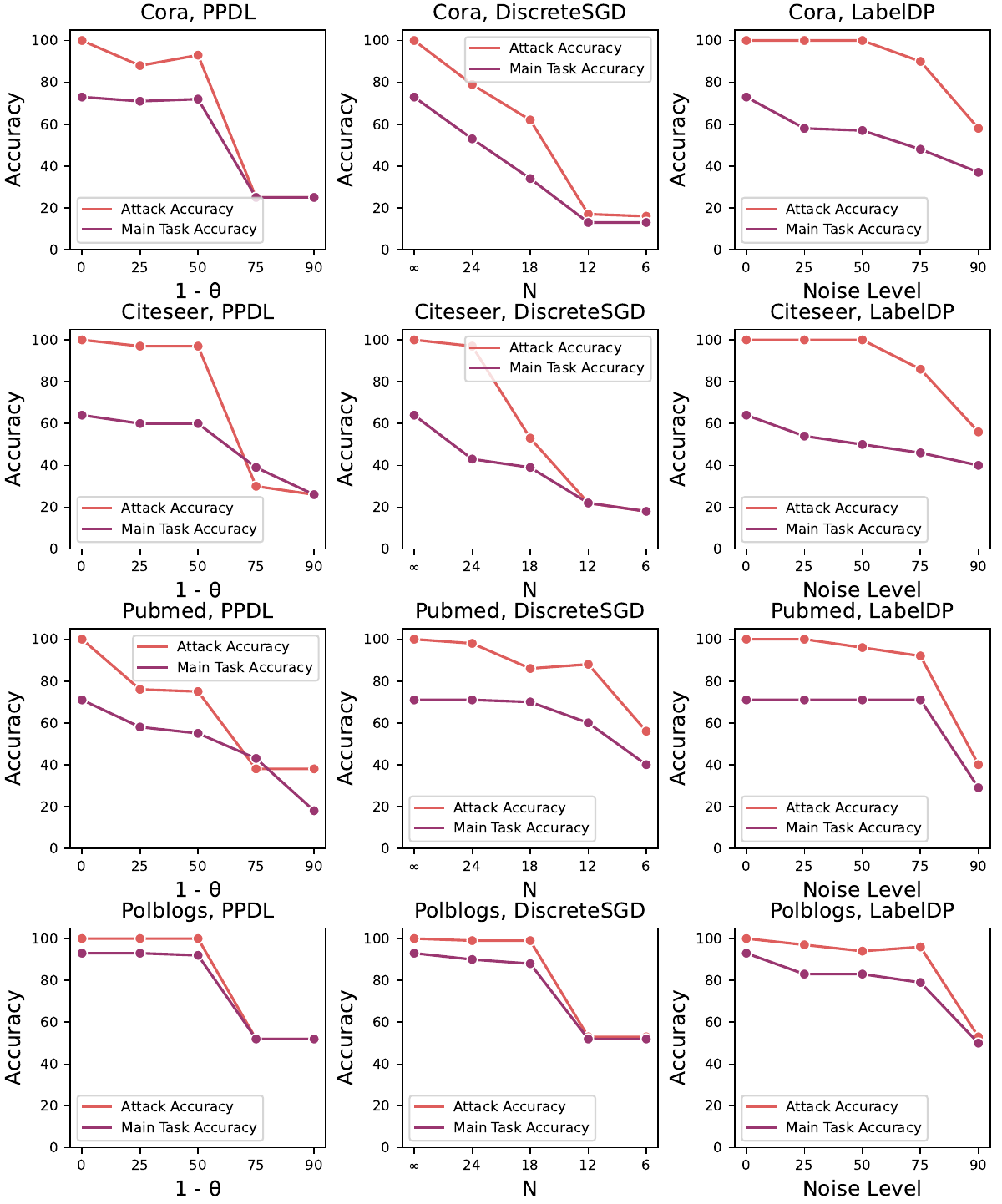}
    \caption{Evaluation of PPDL, DiscreteSGD, and LabelDP defense methods against our proposed attack on four datasets.}
    \label{fig:possible_defense_1}
\end{figure}

We evaluated the six mentioned defenses against our attack using the GCN model on the Cora, Citeseer, Pubmed, and Polblogs datasets against which the attack performed the best. We added Laplacian noise in the second defense method as there is no apparent difference in the performance of the defense method between using Gaussian or Laplacian noise, and both are commonly used in literature~\cite{liu2022federated, qi2020privacy}.
In this evaluation, we tested how the performance of the model on the main task and the accuracy of the attack would change according to the strength of the defenses.
In our tests, we considered the following configurations of the defense methods: clipping threshold $\in \{10^{-1}, 10^{-2}, 10^{-3},10^{-4},10^{-5},10^{-6}\}$, Laplacian noise level $\in \{10^{-1}, 10^{-2}, 10^{-3},10^{-4},10^{-5},10^{-6}\}$, gradient compression percentage $\in \{75\%, 50\%, 25\%, 10\%\}$, PPDL $\theta$ fraction $\in \{10\%, 25\%, 50\%, 75\% \}$, DisctreteSGD number of intervals N $\in \{6, 12, 18, 24\}$, and percentage of probability redistribution on secondary labels $\in \{25\%, 50\%, 75\%, 90\%\}$.
The clipping threshold and noise addition performed similarly. Thus, for better readability and due to the page limit, we report the results on the PPDL, DiscreteSGD, and LabelDP in Figure~\ref{fig:possible_defense_1}, while the results with noise, clipping, and compression are visible in Figure~\ref{fig:possible_defense_2} of Appendix~\ref{sec:noise-defense}.
From the results, we can see that the defense methods must add a large amount of noise, set a small threshold for clipping the gradients, or define a few sub-intervals in the case of DiscreteSGD to be effective. However, such edge configurations significantly impact the performance of the main task.
On the other hand, the Gradient Compression technique, in Figure~\ref{fig:possible_defense_2}, does not degrade the performance of the main task but does not impair the attack's success.
In most cases, the attack's performance is affected when the compression is over 50\%, but even in that case, most of the performance is preserved. 
In the worst case, with compression of 90\%, we can still obtain satisfactory results of over 70\% accuracy for all the datasets. 
To mitigate the attack effectively using described defense methods, one has to sacrifice the performance on the main classification task.
A successful defense method would achieve a more significant decrease in accuracy for the attack than the main task, which was rarely obtained in our experiments. Specifically, the only such case uses a clipping threshold on the Citeseer dataset. The attack accuracy dropped to around $40\%$ while the main task accuracy remained around the initial $60\%$ for the threshold value of $10^{-5}$. 
Thus, we conclude that without compromising the main task accuracy, tested defense methods are ineffective against our proposed attack.
However, starting from the limitation of the analyzed defenses, we can provide a possible direction for the definition of an effective protection that preserves the accuracy of the model. Looking at the ablation study, our attack is highly dependent on the accurate optimization of the server model approximation $g'$. To prevent the attacker from optimizing $g'$, the server can decouple the optimization of the top model $g$ from the clients' bottom model. 
The idea is to calculate the gradients of the bottom model using the same strategy adopted by LabelDP but to include a subsequent additional optimization step using the real labels in the top model. 
We tested this defense proposal and reported preliminary results in Table~\ref{tab:decoupling_defense}. We can see that the defense preserves most of the accuracy of the model on the different datasets, decreasing the attack accuracy $26\%$ on average. Encouraged by these results, as future work, we believe it would be possible to refine LabelDP so to create an obfuscation strategy that is more compatible with the additional optimization strategy available only on the top model. In this way, we might be able to push more on the label obfuscation while still being able to reconstruct the top model accuracy in the subsequent step.

\begin{table}[]
\centering
\caption{Evaluation of the proposed possible defense against our attack on four datasets.}
\label{tab:decoupling_defense}
\resizebox{\columnwidth}{!}{%
\begin{tabular}{ccccc}
\hline
Dataset  & \begin{tabular}[c]{@{}c@{}}Main Task Accuracy\\ Without Defense\end{tabular} & \begin{tabular}[c]{@{}c@{}}Attack Accuracy\\ Without Defense\end{tabular} & \begin{tabular}[c]{@{}c@{}}Main Task Accuracy\\ With Defense\end{tabular} & \begin{tabular}[c]{@{}c@{}}Attack Accuracy\\ With Defense\end{tabular} \\ \hline
CORA     & 72\%                                                                         & 100\%                                                                     & 72\%                                                                      & 74\%                                                                   \\
Citeseer & 64\%                                                                         & 100\%                                                                     & 60\%                                                                      & 73\%                                                                   \\
Pubmed   & 71\%                                                                         & 100\%                                                                     & 71\%                                                                      & 76\%                                                                   \\
Polblogs & 93\%                                                                         & 100\%                                                                     & 84\%                                                                      & 72\%                                                                   \\ \hline
\end{tabular}%
}
\end{table}

\section{Related Work}
\label{sec:related work}

This section discusses related works on federated learning with GNNs, followed by an overview of attacks in federated learning.
Federated Learning (FL) has seen widespread applications across various domains, but its adoption for graph data remains relatively limited. Most research works focus on Horizontal FL (HFL) and present several HFL frameworks tailored to GNN architectures, such as ASFGNN~\cite{zheng2021asfgnn}, GraphFL~\cite{wang2022graphfl}, and FedGraphNN~\cite{he2021fedgraphnn}.
In contrast, GNN-based Vertical FL has only recently gained attention. Chen et al.~\cite{chen2020vertically} introduced the first paradigm for privacy-preserving node classification in GNN-based VFL, named Vertically Federated Graph Neural Network (VFGNN). To ensure privacy, the authors proposed incorporating differential privacy mechanisms to prevent potential information leakage from the server.
Another notable framework in the context of privacy-preserving node classification is FedVGCN, introduced by Ni et al.~\cite{ni2021vertical}. This framework extends to Graph Convolutional Networks (GCN). Like VFGNN, the training process involves two parties that exchange intermediate results during each training iteration, but, in this case, the values are exchanged under Homomorphic Encryption (HE)~\cite{a-fully-homomorphic-encryption-scheme}.
In this paper, we considered a classical vertical federated framework involving active and passive clients. Using the default framework for this task aims to reproduce the most realistic setting, putting the attacker in the worst-case scenario. In particular, we considered standard and realistic architectures that do not give any advantage to the attacker, who can only exploit the information leaked by the gradients. Considering Horizontal FL, Hitaj et al.~\cite{hitaj2017deep} used generative adversarial networks to generate the same distribution as the training data. Nasr et al.~\cite{nasr2019comprehensive} exploited the stochastic gradient descent (SGD) vulnerability to develop a membership attack in a setting where the adversary has complete knowledge of the model's architecture and parameters. 
Zhu et al.~\cite{zhu2019deep} used public shared gradients to obtain the local training data and labels without extra knowledge.
The malicious party tries to match the real gradients with what they call dummy gradients derived from randomly initialized synthetic data and labels. 
Our work uses a similar matching gradients process, but the framework is different and does not use synthetic data. 
Zhao et al.~\cite{zhao2020idlg} demonstrated that the ground truth label could be extracted by leveraging the directions of the gradients of the weights connected to the logits of different classes.
Most of these attacks and exploits rely on the gradient exchanged during training. 
However, these attacks work under different settings, horizontal data partitioning, and none consider GNN models. Moreover, the goals of the described attacks differ from ours. In vertical setup, Luo et al.~\cite{luo2021feature} formulated a problem of feature inference attack, and they proposed several attack methods on logistic regression and decision trees. They also offered a generalization of the attack to be applied to neural networks or random forest models. Li et al.~\cite{li2021labelleakgesplitlearning} explored whether the labels can be uncovered by sharing gradient in the two-party split learning with feature-partitioned data. Fu et al.~\cite{fu2022label} proposed passive, active, and direct label inference attacks. The passive attack requires a small amount of label data, while our approach does not.
An active attack defines a specialized malicious local optimizer that makes the federated model rely more on the malicious local model than other parties. Lastly, the direct label inference attack relies on gradient signs to deduce the correct label. However, the gradients must be derived directly from the final prediction layer. In particular, the authors assume that the client's bottom models produce embeddings with dimensions equal to the number of classes, and the server, instead of having a top model, calculates the loss by comparing the embeddings directly with the corresponding one-hot encoded labels.
Our approach, instead, can be effective in any configuration of the VFL framework.
Zou et al.~\cite{zou2022defending} presented an analysis of a batch-level label inference attack within the VFL setting with HE protection that can recover the labels when the batch size is smaller than the dimension of the embeddings of the last connected layer. They used the batch-level gradients of the local model generated using the one associated with each sample in the considered batch to leak the corresponding labels and the representation of the nods of the other clients.
However, instead of using the batch-level gradients, we exploit the gradients of the local model computed starting from all of the training samples. Instead of trying to infer the embeddings of the other clients, we replicate the entire stack of training approximating a server model. This additional step helps to mitigate possible defense strategies since our attack uses only the gradients generated by the manipulated clients for its specific local model instead of trying to match the privacy-enforced partial gradients returned by the server. There are limited studies on adversarial attacks in GNN-based VFL frameworks. For instance, Chen et al.~\cite{chen2022graphfraudster} presented an attack that aims to mislead the targeted GNN model into producing a desired label for a given adversarial example. The attacker achieves this by stealing node embeddings and introducing noise to confuse the server's shadow model.
Another study focused on the leakage of sample relations in a GNN-based VFL setting, which should maintain private~\cite{qiu2022yourlabelsselling}. The authors proposed three types of attacks based on different intermediate representations, depending on the adversary's level of knowledge.
It is important to note that our work does not directly compare with the mentioned attacks, as they have distinct malicious objectives compared to ours, even though they are conducted in a similar setting using GNN-based VFL. We focus on the novel label inference attack, unveiling new potential vulnerabilities in GNN-based VFL. To our knowledge, our work is the first to introduce a label inference attack explicitly targeting the GNN-based VFL scenario and propose this attack in the VFL setting using no additional knowledge.

\section{Conclusions and Future Work}
\label{sec:conclusion}

Vertical federated learning is a privacy-preserving collaborative learning paradigm for training machine learning models. There, parties have the same data sample space, but the local private data differ in feature space.  
In a typical VFL scenario, the labels of the samples are considered private information that should be kept secret from all the parties (passive users) except for the aggregating server (active user). Intermediate gradient values are exchanged and, in the past, assumed safe to share. Recent works have shown that the standard VFL setup, even using encryption, leaks sensitive information such as training data or labels.
Several works showcase label inference attacks on the VFL setup. 
In this paper, we are the first to test a label inference attack on graph neural networks for the node classification task in a VFL framework, using no background knowledge of the labels. 
We tested our attack, called BlindSage, on different GNN models, datasets, and attackers' knowledge assumptions. 
Based on the experiments, the BlindSage attack achieves $100\%$ accuracy in most cases. 
We also propose an early stopping strategy for a more efficient solution and analyze the role of specific attack components. We found that a refined server model approximation is the most significant factor for the attack's success. 
Lastly, we tested the BlindSage attack against defenses such as clipping threshold, noise addition, and gradient compression, showing BlindSage is robust against them. 
We identified a promising strategy to build an enhanced defense against our attack. Preliminary results showed it could be effective in reducing the attack success rate without impacting the federated model performance.
Thus, in future work, we will further explore the proposed defense strategy to provide enhanced protection for the GNN-based VFL framework. 
Moreover, we plan to further develop our proposed label inference attack by exploiting some graph-specific characteristics to improve the attack's efficiency and investigate how well it works on different neural networks, such as CNNs or transformer networks.








%



\appendix

\subsection{Architectures}
\label{sec:architectures}

In our experiments, we focused on three standard GNN architectures: GCN~\cite{kipf2017semigcn}, GAT~\cite{velickovic2018gat}, and GraphSAGE~\cite{hamilton2017inductiverepr_graphsage_reddit}. 
As for the hyperparameters configuration of the corresponding models, we adopted the standard settings declared by previous works in the scientific literature. 
Specifically, for the GCN model as done in~\cite{kipf2017semigcn}, we used $2$ aggregation layers with the hidden dimension of $32$ and a ReLU activation function between them.
As for the GAT model, we followed the configuration described in~\cite{velickovic2018gat}, i.e., we used $3$ aggregation layers with $3$ heads each.
For the GraphSAGE model, as commonly done, we used $2$ aggregation layers with a hidden dimension of $32$. Furthermore, in the GraphSAGE model, we apply the MEAN aggregator, as mentioned in Section~\ref{gnn}.
We opted for the most representative types of GNNs. Indeed, more complex ones adding numerous layers to GNNs negligibly improve performance, and sometimes, they even harm node classification~\cite{oono2019graph, alon2020bottleneck}.

All the experiments have been carried out using an AMD Ryzen 5800X CPU paired with 32 GB of RAM and a Tesla T4 GPU with 16GB of VRAM. Our label inference attack and VFL have been developed using PyTorch 2.0.


  

\subsection{Datasets}
\label{sec:datasets}

To carry out a complete study of our attack's behavior, we selected datasets with different complexity in terms of the number of classes, edges, and nodes.
The most common datasets considered in the related literature about vertical federated learning and GNNs~\cite{chen2022graphfraudster,chen2020vertically,ni2021vertical,qiu2022yourlabelsselling} are Cora~\cite{mccallum2000automating_corads}, Citeseer~\cite{giles1998citeseer}, and Pubmed~\cite{sen2008collective}. Additionally, we considered Polblogs~\cite{adamic2005politicalblogosphere}, which has only two classes and fewer nodes.
On the other hand, we try our method on the arXiv~\cite{hu2021opengraphbenchmark} dataset, which has a much larger number of nodes, edges, and classes. We also consider the Reddit~\cite{hamilton2017inductiverepr_graphsage_reddit} dataset as it represents social network data.
Finally, we also considered the Reddit~\cite{hamilton2017inductiverepr_graphsage_reddit} and Arxiv~\cite{hu2021opengraphbenchmark} datasets, which have a large number of nodes, edges, and classes. Moreover, the Reddit dataset appears particularly interesting as it allows us to test our strategy in a different yet extremely popular context regarding privacy concerns, i.e., social networking.
With the selected pool of datasets, we aim to offer results that can be directly comparable for benchmarking purposes while ensuring enough data diversity to demonstrate our attack's general applicability further.
In Table~\ref{tab:datasets}, we report the basic information on the datasets, e.g., the number of nodes and edges. The last column in this table lists previous related works on VFL and GNN that used the given dataset. 
\begin{table}[!ht]
\centering
\caption{Information on utilized datasets.}
\label{tab:datasets}
\resizebox{\columnwidth}{!}{%
\begin{tabular}{@{}l *{4}{c}r@{}}
\toprule
 Dataset & Nodes & Edges & Features & Classes & Used in (only VFL  \\ 
         &       &       &          &         &   GNN papers)      \\ \midrule
Cora~\cite{mccallum2000automating_corads} & 2\,708 & 5\,278   & 1\,433       & 7         & \cite{chen2022graphfraudster,chen2020vertically,ni2021vertical,qiu2022yourlabelsselling} \\
Citeseer~\cite{giles1998citeseer}   & 3\,327 & 9\,228 & 3\,703       & 6         &  \cite{chen2022graphfraudster,chen2020vertically,ni2021vertical,qiu2022yourlabelsselling}\\
Pubmed~\cite{sen2008collective}    & 19\,717 & 88\,651 & 500        & 3         & \cite{chen2022graphfraudster,chen2020vertically,ni2021vertical} \\
Polblogs~\cite{adamic2005politicalblogosphere} & 1\,222 & 33\,431                 & 1\,222          & 2         & \cite{chen2022graphfraudster} \\
arXiv~\cite{hu2021opengraphbenchmark}      & 169343 & 2315598              & 128        & 40        & \cite{chen2020vertically} \\
Reddit~\cite{hamilton2017inductiverepr_graphsage_reddit}     & 232\,965      & 57\,307\,946               & 602        & 41        &  / \\ \bottomrule
\end{tabular}%
}
\end{table}

\subsection{Analysis of the Number of Attack Iterations}
\label{sec: attack iterations analysis}

In all our experiments, we set the number of iterations for the attack during the training of the main task, $AttackIterations$, to $10$. 
Here, we report the results that allowed us to identify the most promising setting for such value.
To carry out this analysis, we tested different values for the number of attack iterations, namely $\{5, 10, 15\}$.
To keep the experiment practical from a computation point of view, we considered a smaller subset of the GNN architectures and datasets.
In particular, we focused on the GCN and GraphSAGE architectures using the Cora and Citeseer datasets.
The results are reported in Table~\ref{tab:diff attack iterations}, in which the ones with the best accuracy are highlighted in bold. 
As visible from this table, in most cases, the best results are obtained with $10$ attack iterations. In the most challenging scenario, i.e., the {\em no knowledge} case, for configuration with the GCN model, $15$ iterations were slightly better to maximize the obtained accuracy. 
However, except for this case, a setting with $10$ iterations also appears more suitable because the best accuracy is obtained after fewer training epochs. 

\begin{table}[!ht]
\small
\centering
\caption{Attack accuracy on two GNN models and two common graph datasets with different numbers of $AttackIterations$ and the optimal learning rate (lr).}
\label{tab:diff attack iterations}
\resizebox{\columnwidth}{!}{%
\begin{tabular}{@{}cccccc@{}}
\toprule
&    &   & \multicolumn{1}{l}{Basic Knowledge} & Limited Knowledge & \multicolumn{1}{l}{No knowledge} \\ 
Architecture & Dataset & Attack iter. & lr=0.1 & lr=0.5 & lr=1 \\ \midrule
\multirow{6}{*}{GCN}       & \multirow{3}{*}{CORA} & 5 & 98\%                               & 100\%             & 92\%                             \\
 &                           & 10                 & \textbf{100\%}           & 100\%                  & 93\%          \\
 &                           & 15                 & 100\%                    & 100\%                  & \textbf{94\%} \\
 & \multirow{3}{*}{Citeseer} & 5                  & 100\%                    & 99\% (lr=1)            & 83\%          \\
 &                           & 10                 & 100\%                    & \textbf{100\%}  (lr=1) & 87\%          \\
 &                           & 15                 & 100\%                    & 100\%  (lr=1)          & \textbf{91\%} \\ \midrule
\multirow{6}{*}{GraphSAGE} & \multirow{3}{*}{CORA} & 5 & 99\% (lr=0.5)                      & 100\%             & 96\%                             \\
 &                           & 10                 & \textbf{100\%}  (lr=0.5) & 100\%                  & \textbf{98\%} \\
 &                           & 15                 & 100\%  (lr=0.5)          & 100\%                  & 97\%          \\
 & \multirow{3}{*}{Citeseer} & 5                  & 98\%                     & 100\%                  & 98\%          \\
 &                           & 10                 & \textbf{100\%}           & 100\%                  & 98\%          \\
 &                           & 15                 & 100\%                    & 100\%                  & 98\%          \\ \bottomrule
\end{tabular}%
}
\end{table}

\subsection{Early Stopping Strategy for BlindSage: GAT Architecture}
\label{sec:early-stopping-gat}

\begin{figure*}[!ht]
    \centering
    \includegraphics[width=0.86\textwidth]{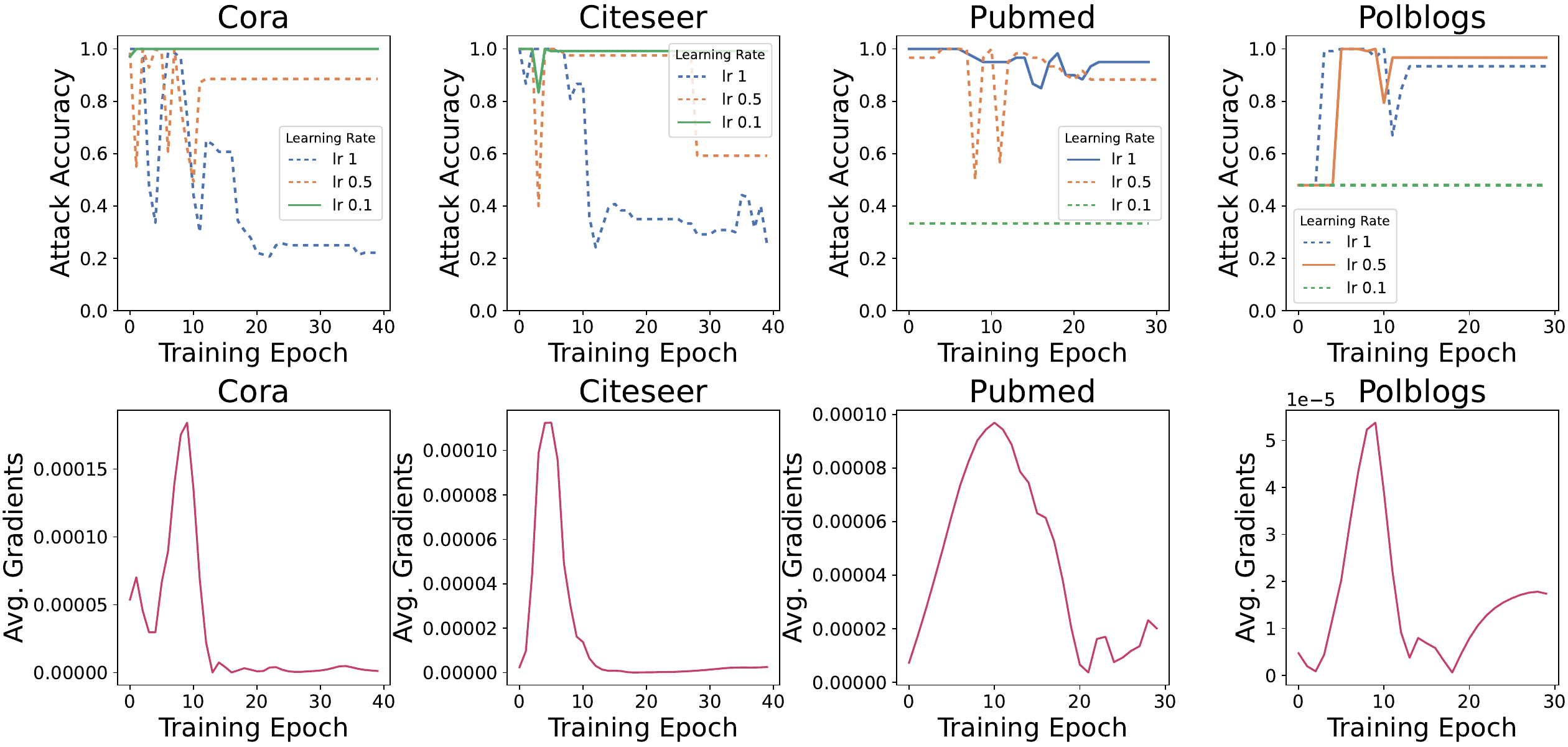}
    \caption{Attack accuracy versus the average values of the local model gradients in a {\em basic knowledge} scenario using the GAT architecture.}
    \label{fig:grad_accuracy_gat}
\end{figure*}

In Figure~\ref{fig:grad_accuracy_gat}, we show in the first row the accuracy of our attack during the training of the VFL model using the GAT architecture as the local model, and in the second row, the average of the gradients associated with each epoch. 

\subsection{Potential Defenses: Noise Addition}
\label{sec:noise-defense}

\begin{figure*}
    \centering
    \includegraphics[width=0.86\textwidth]{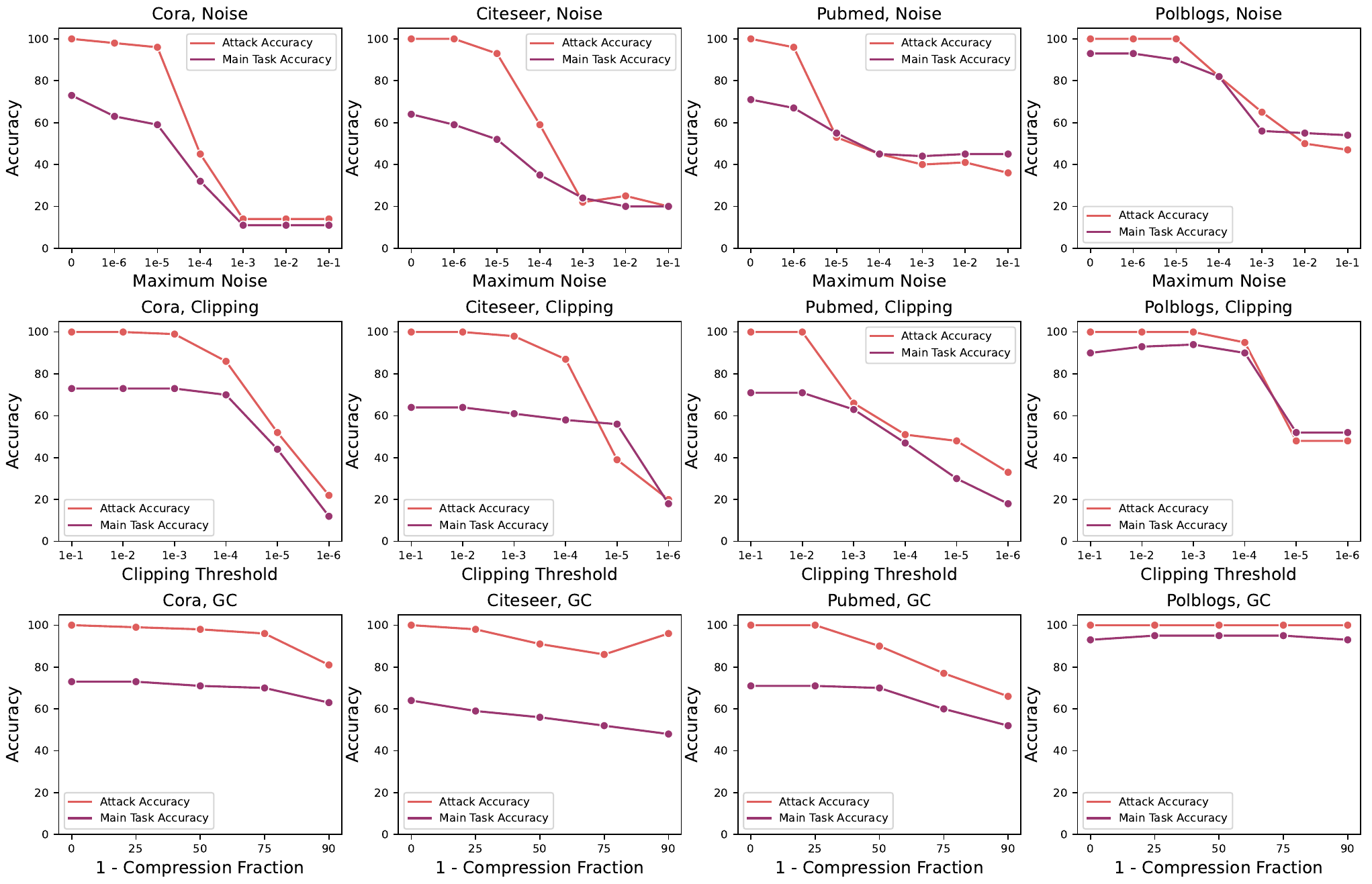}
    \caption{Evaluation of the noise addition defense method against our proposed attack on four datasets. \label{fig:possible_defense_2}}
\end{figure*}

In Figure~\ref{fig:possible_defense_2}, we show the results obtained using the Laplacian noise addition, clipping, and gradient compression as defenses against the BlindSage attack. The results are similar to the performance of the other methods (see Figure~\ref{fig:possible_defense_1} in Section~\ref{sec:defense}).

\subsection{Test on Different Combination and Aggregation Techniques}
\label{sec:combinaton-aggregation}

In Table~\ref{tab:combinaton-aggregation}, we show the results obtained against the different combinations of three aggregators (two basic ones, i.e., ``max'' and ``mean'', and a more advanced one, namely ``std'') and the combination techniques MEAN and CONCAT. For this experiment, we chose the Cora, Pubmed, and Citeseer datasets as test benches. It is worth noting that since the attacker has full control of its local model, the attacker can always select the most promising aggregation technique. The important result of this experiment is that our attack is agnostic to the embedding combination technique employed by the server.

\begin{table*}[!ht]
\caption{Evaluation of the attack against different combinations of combination and aggregation techniques for the GraphSage architecture.}
\label{tab:combinaton-aggregation}
\resizebox{\textwidth}{!}{%
\begin{tabular}{ccccccc|cccccc|cccccc}
\hline
\multicolumn{1}{l}{} & \multicolumn{6}{c|}{CORA}                              & \multicolumn{6}{c|}{Pubmed}                            & \multicolumn{6}{c}{Citeseer}                          \\ \hline
                     & \multicolumn{3}{c}{CONCAT} & \multicolumn{3}{c|}{MEAN} & \multicolumn{3}{c}{CONCAT} & \multicolumn{3}{c|}{MEAN} & \multicolumn{3}{c}{CONCAT} & \multicolumn{3}{c}{MEAN} \\ \hline
                     & MAX    & MEAN    & STD     & MAX     & MEAN   & STD    & MAX     & MEAN    & STD    & MAX     & MEAN   & STD    & MAX    & MEAN    & STD     & MAX    & MEAN   & STD    \\ \hline
Basic Knowledge      & 99\%   & 100\%   & 100\%   & 100\%   & 100\%  & 100\%  & 95\%    & 90\%    & 88\%   & 96\%    & 90\%   & 95\%   & 98\%   & 100\%   & 100\%   & 99\%   & 100\%  & 100\%  \\
Limited Knowledge    & 97\%   & 96\%    & 98\%    & 97\%    & 98\%   & 97\%   & 93\%    & 90\%    & 88\%   & 93\%    & 90\%   & 95\%   & 95\%   & 100\%   & 100\%   & 99\%   & 100\%  & 100\%  \\
No Knowledge         & 93\%   & 94\%    & 98\%    & 93\%    & 97\%   & 97\%   & 91\%    & 86\%    & 83\%   & 93\%    & 86\%   & 91\%   & 85\%   & 98\%    & 95\%    & 80\%   & 96\%   & 98\%   \\ \hline
\end{tabular}%
}
\end{table*}

\end{document}